\documentclass{article}

\usepackage{arxiv}

\usepackage[utf8]{inputenc} 
\usepackage[T1]{fontenc}    
\usepackage{hyperref}       
\usepackage{url}            
\usepackage{booktabs}       
\usepackage{amsfonts}       
\usepackage{nicefrac}       
\usepackage{microtype}      
\usepackage{lipsum}
\usepackage{graphicx}
\usepackage{multirow}%
\usepackage{amsmath,amssymb,amsfonts}%
\usepackage{amsthm}%
\usepackage{mathrsfs}%
\usepackage[title]{appendix}%
\usepackage{xcolor}%
\usepackage{textcomp}%
\usepackage{manyfoot}%
\usepackage{algorithm}%
\usepackage{algorithmicx}%
\usepackage{algpseudocode}%
\usepackage{listings}%
\graphicspath{ {./images/} }

\title{Unsupervised Keypoints for Real-Time Fall Detection: Comparative Analysis Under Real-world Conditions with Predictive Bandwidth Reduction}

\author{
 Tasmiah Haque \\
  Department of Industrial and Management Systems Engineering\\
  West Virginia University\\
  Morgantown, WV 26505 \\
  \texttt{th00027@mix.wvu.edu} \\
   \And
 Jacob Kosinski \\
  Lane Department of Computer Science and Electrical Engineering\\
  West Virginia University\\
  Morgantown, WV 26505 \\
  \texttt{jk00067@mix.wvu.edu} \\
  \And
 Sumit Mohan \\
  Intel Corporation\\
  Santa Clara, CA 95054 \\
  \texttt{sumit.mohan@intel.com} \\
\And
  Mohammad Abdullah Al-Mamun \thanks{Corresponding author}\\
  School of Systems Science and Industrial Engineering\\
  Binghamton University\\
  Binghamton, NY 13902\\
  \texttt{malmamun1@binghamton.edu} \\
\And
  Srinjoy Das \footnotemark[1] \\
  School of Mathematical and Data Sciences\\
  Department of Industrial and Management Systems Engineering\\
  West Virginia University\\
  Morgantown, WV 26505 \\
  \texttt{srinjoy.das@mail.wvu.edu} \\
}

\begin{document}
\maketitle
\begin{abstract}
Falls among older adults are a major safety and health-systems challenge, yet continuous in-person monitoring is difficult to sustain across home and clinical care settings. Video-based monitoring can capture fall-relevant motion, but scalable real-time deployment is limited by privacy, compute, and bandwidth constraints, and existing keypoint-based methods typically rely on supervised or anatomical pose representation, which is vulnerable to occlusion and partial body visibility. We propose a fall-monitoring framework that replaces continuous video transmission with compact motion representations, using unsupervised keypoints which are extracted locally, and a variational recurrent model is used to forecast motion at the staff end, followed by fall classification. We evaluate the framework on the UR Fall and Human Fall datasets under random, subject-disjoint, and occlusion-based splits to systematically characterize when each representation has an advantage. We find that random splits do not discriminate between representations, and under subject-disjoint evaluation no uniform advantage emerges; performance varies across held-out subjects with their visual characteristics. Under occlusion, however, unsupervised keypoints substantially outperform supervised keypoints, retaining strong detection sensitivity where supervised keypoints miss approximately half of falls; this advantage reflects their anatomical independence and persists under bandwidth-constrained prediction. The unsupervised detector also requires roughly two orders of magnitude less computation, supporting privacy-preserving, bandwidth-aware, always-on fall monitoring. The code is available at: https://github.com/Tasmiah1408028/Unsupervised-Keypoints-for-Real-Time-Fall-Detection
\end{abstract}


\section{Introduction}

Among adults aged 65 years and older, falls are the leading cause of both fatal and nonfatal injury, and their burden rises sharply with age. In the United States nearly 1 in 4 adults aged 65 years and older experiences a fall each year, amounting to approximately 14 million older adults annually and resulting in nearly 3 million emergency department visits \cite{NCOA2025FallsFacts}. In hospitals alone, an estimated 700,000 to 1 million falls occur annually, corresponding to 3 to 5 falls per 1,000 patient-days \cite{NCOA2025FallsFacts, Osonuga2026AIHospitalFallPrevention}. Falls also contribute substantially to mortality, with approximately 38,000 deaths among older adults reported in 2021 \cite{NCOA2025FallsFacts}.
 
Fall-related injuries can lead to hip fractures \cite{wiklund2016risk}, head trauma \cite{luukinen2005fall}, fear of falling \cite{hoang2017factors}, reduced mobility, loss of independence, and nearly doubled hospital stays \cite{NCOA2025FallsFacts}; among older adults, serious falls are also associated with substantial mortality, with post-injury mortality reported at roughly 10--15\% \cite{Osonuga2026AIHospitalFallPrevention}. A particularly dangerous consequence is prolonged time spent on the ground after the fall, since extended immobilization has been linked to dehydration, pressure injury, rhabdomyolysis, and pneumonia \cite{Kubitza2023LongLie, Blackburn2022LongLieSystematicReview}, making rapid time-to-awareness and response clinically important.
 
The economic burden of falls is correspondingly large. Non-fatal falls among older adults account for an estimated \$80 billion in annual healthcare costs, and this figure is projected to rise to \$101 billion by 2030 \cite{Houry2016CDCFalls, Haddad2024HealthcareSpendingFalls}, with Medicare bearing approximately 67\% of these expenditures \cite{Haddad2024HealthcareSpendingFalls}. Shortening the delay to intervention therefore has importance not only for patient safety, but also for reducing a substantial Medicare-related expenditure burden. These statistics underscore that falls are not isolated incidents but rather a persistent source of morbidity, mortality, functional decline, and avoidable healthcare spending in aging populations. In the broader context of rising Medicare expenditures, fall prevention and timely post-fall response are therefore important not only from a patient-safety perspective but also from a health-systems and reimbursement perspective. Technologies that can shorten time spent unattended after a fall may help address both clinical and economic costs.
 
These needs are especially pressing because most falls among older adults occur in the home rather than in continuously observed care settings; one review reported that more than 70\% of community falls occur at home \cite{soriano2007falls}. This makes fall-focused remote patient monitoring particularly relevant to aging-in-place and rural care delivery, where older adults may live far from hospitals. Telehealth often serves as the main bridge to clinicians \cite{FCC2024Section706, Rush2022TelehealthRuralOlderAdults} as continuous in-person observation is rarely feasible. However, conventional remote monitoring approaches remain limited. Wearable devices and call buttons can be useful, but they may be forgotten, declined, removed, or unable to capture contextual information about how an event unfolded \cite{Warrington2021WearableFalls}. By contrast, video-based monitoring can capture rich information about posture, movement, environment, and temporal dynamics, making it attractive for detecting falls and related mobility risks in both home and facility settings \cite{Shaik2023RemotePatientMonitoringAI, Gabriel2025ContinuousPatientMonitoringAI, Habibi2025VideoBasedHOI}. Recent work has also shown that camera-based patient monitoring can extend beyond binary fall alarms to capture clinically relevant behavioral trends, including patient isolation, unsupervised movements, ambulation, and interactions with the environment, thereby supporting broader safety and workflow objectives in hospitals and recovery settings \cite{Gabriel2025ContinuousPatientMonitoringAI, Habibi2025VideoBasedHOI}.
 
Despite this promise, real-time deployment of video-based remote monitoring for older adults remains constrained by three interrelated barriers: privacy, deployment cost, and bandwidth. First, continuous RGB video exposes identity, living conditions, and other highly sensitive contextual information, creating significant privacy and regulatory concerns \cite{Shaik2023RemotePatientMonitoringAI, Mujirishvili2023AcceptancePrivacyVideoAAL}. Second, raw-video systems are costly to deploy because they require sustained computation, storage, networking, and secure data management across long monitoring periods \cite{Shaik2023RemotePatientMonitoringAI}. Third, continuous transmission of video is bandwidth-intensive, which is particularly problematic in rural and underserved regions where broadband access remains limited or inconsistent \cite{FCC2024Section706, Rush2022TelehealthRuralOlderAdults}. These barriers are especially important for rural remote monitoring, where one of the very populations that could benefit most from longitudinal monitoring may also face the greatest connectivity constraints. A practical monitoring system for rural America must therefore be privacy-preserving, lightweight, and able to operate under constrained bandwidth conditions. Even with identity blurring, recent hospital-scale AI video monitoring \cite{Gabriel2025ContinuousPatientMonitoringAI} still faces real-world deployment challenges from continuous streaming, variable room conditions, and the need for scalable real-time inference.
 
These limitations motivate the exploration of unsupervised keypoint learning \cite{jakab2018unsupervised} for fall monitoring that is optimized for motion discrimination, temporally predictive, and compatible with deployment constraints such as privacy, bandwidth, and compute. By learning motion landmarks directly from video without manual pose labels, unsupervised keypoint detectors offer the potential for a lighter, more scalable, and more privacy-aligned front-end \cite{Minderer2019UnsupervisedObjectStructureDynamics}. Such representations may also be better suited to out-of-distribution settings if they capture motion structure rather than depending on fine-grained anatomical localization for practical healthcare deployment. 
 
Our central hypothesis is that, for real-world healthcare fall monitoring, anatomical pose accuracy is not necessarily the right representation objective. Despite the growing body of work on pose-based fall detection, existing literature has paid limited attention to out-of-distribution evaluation. Most studies report performance under controlled benchmark conditions leaving open the question of how these representations behave under the visual degradations commonly encountered in real deployment, such as occlusion, partial body visibility, and subject appearance variation. 
From this perspective, we study whether unsupervised motion keypoints provide a more suitable clinical representation than supervised pose under realistic monitoring conditions, and we systematically characterize the conditions under which each approach has an advantage.
 
\begin{figure}
\centering
\includegraphics[width=\textwidth]{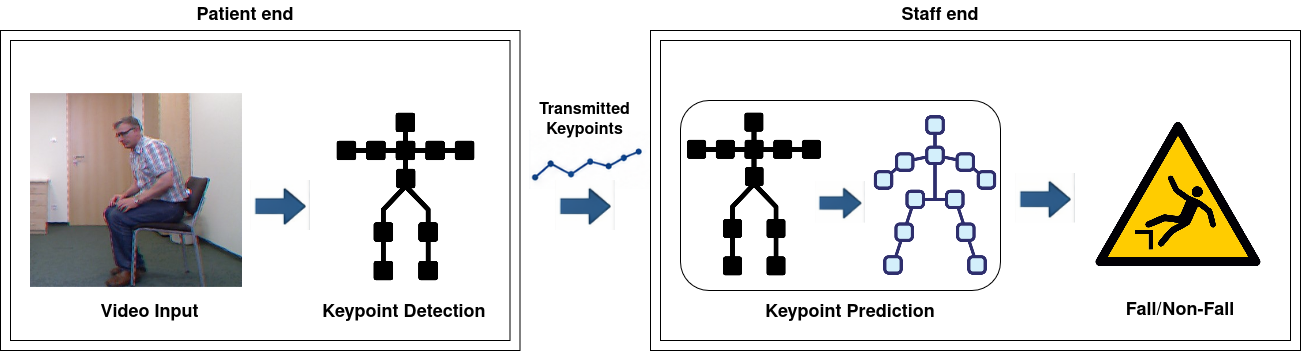}
\caption{Overview of our proposed privacy-preserving remote fall-monitoring pipeline}\label{fig_1}
\end{figure}

In this work, we investigate a privacy-preserving remote fall-monitoring pipeline built around unsupervised motion keypoints and predictive temporal modeling, as shown in Figure~\ref{fig_1}. We adopt a distributed edge-cloud architecture, where rather than transmitting continuous video, our approach extracts compact motion keypoints locally at the patient home or aging-in-place facility using an unsupervised keypoint predictor. Based on these transmitted keypoints, a recurrent generative model is then used to forecast future motion at the staff end at a centralized clinical monitoring system, followed by fall/non-fall classification from the resulting motion representation.

This design directly addresses the three core deployment constraints motivating the study. Privacy is improved because identity-rich imagery need not be transmitted continuously; deployment cost is reduced by using a lightweight motion representation rather than a heavy raw-video pipeline; and bandwidth requirements are lowered because compact keypoints, rather than full video streams, can be communicated in real time \cite{Shaik2023RemotePatientMonitoringAI, Mujirishvili2023AcceptancePrivacyVideoAAL}. The keypoint predictor at the clinical monitoring system, realized via the recurrent generative model, provides additional bandwidth savings by enabling intermittent transmission of observed keypoints, with intermediate motion inferred through prediction rather than continuous streaming. By explicitly focusing on real-world conditions and bandwidth-constrained monitoring, this work targets the conditions most relevant to rural remote care and real-world clinical deployment rather than only favorable benchmark settings. To the best of our knowledge, this is the first work to apply unsupervised keypoint learning i.e. keypoints learned from video without manual joint annotations to fall detection under real-world conditions.
 
We make the following contributions:
\begin{enumerate}
    \item We show that anatomically supervised pose is not necessarily the optimal motion representation for fall detection under realistic deployment constraints, and we identify the specific conditions under which each approach (supervised/unsupervised) has an advantage.
    \item We introduce a privacy-preserving remote fall-monitoring pipeline based on unsupervised motion keypoints, predictive temporal modeling, and lightweight sequence classification.
    \item We demonstrate that unsupervised keypoints provide advantages over supervised keypoints specifically under occlusion conditions, due to their anatomical independence, while the supervised representation tends to be favored under clean full-body visibility.
    \item We characterize the accuracy and deployment cost of predictive bandwidth reduction, in which only half of the keypoint frames are transmitted and the remainder are forecast at the receiver, and we discuss how the same predictive mechanism provides a defined fallback when keypoints are lost in transmission.
\end{enumerate}

\section{Related Work}
 
\subsection{Fall Detection Approaches}
 
A large body of prior fall-detection literature has explored wearable, multimodal, RGB, depth, thermal, and skeleton-based approaches. Wearable and inertial systems can be effective, but they depend on user compliance and may be inconvenient for older adults, especially those with cognitive impairment. Multimodal sensor-fusion approaches have shown strong classification performance by combining RGB and inertial signals, but they also increase sensing and system complexity \cite{Abro2025MultimodalBiosensorsFall}.
 
In the vision domain, researchers have pursued both raw-RGB models and privacy-preserving higher-level motion abstractions \cite{NunezMarcos2024TransformerFallDetection, Nogas2020DeepFall, Shin2025ThreeStreamGCNFall} using transformers \cite{vaswani2017attention} and convolutional networks. RGB methods based on transformers have demonstrated that fall detection can be performed directly from video frames without requiring handcrafted auxiliary representations \cite{NunezMarcos2024TransformerFallDetection}, while other work has framed fall detection as anomaly detection using spatio-temporal autoencoders trained on normal activities \cite{Nogas2020DeepFall, Shin2025ThreeStreamGCNFall}. However, anomaly-detection formulations learn normal behavior rather than fall-discriminative motion patterns, which can reduce performance when normal activities resemble falls in individual frames. More broadly, raw-RGB approaches operate in a rich appearance space and are therefore more computationally demanding than compact motion representations, which limits their suitability for bandwidth-constrained and privacy-sensitive deployment.
 
\subsection{Keypoint and Pose-Based Fall Detection}
 
One way to address these limitations is to replace raw visual data with compact motion representations. Keypoint-based or pose-based representations; typically 2D skeletal joint coordinates extracted from videos \cite{chen20232d, salisu2023survey} are appealing because they preserve clinically relevant information about posture and motion while discarding much of the appearance detail that makes RGB video privacy-sensitive. Recent studies have combined human pose estimation with neural network classifiers to build lightweight, privacy-preserving fall-detection systems that can operate locally on low-power devices and transmit only alerts rather than raw imagery \cite{Sykes2024NextGenerationFall, asif2020privacy, liang2024skeleton, yu2025real, analia2025privacy, lin2020framework}.
 
Typically such pipelines rely on supervised keypoint or pose detectors trained on manually annotated body-joint labels \cite{Cao2021OpenPose, kendall2015posenet, google2024mediapipe, Wu2026HEViTPose}. Although such systems can perform strongly under favorable benchmark conditions, they carry several limitations that are problematic for real-world healthcare deployment. They depend on costly labeled pose datasets, often require favorable imaging conditions for reliable joint localization, and can introduce a heavier computational front-end into the monitoring pipeline. More fundamentally, these limitations raise a representation question: for fall detection in healthcare settings, is anatomically correct pose estimation the right objective at all? Falls are distinguished not only by joint locations, but by the temporal evolution of posture and motion under noisy, heterogeneous, and deployment-constrained conditions.

\subsection{Unsupervised Keypoint Learning}
 
Unsupervised keypoint detectors learn compact spatial representations directly from unlabeled images or video, without requiring manual joint annotations \cite{jakab2018unsupervised}. Early work demonstrated that keypoint-like structures emerge from image reconstruction objectives when combined with appropriate spatial bottleneck constraints \cite{jakab2018unsupervised}. Subsequent work extended this to dynamic settings by coupling keypoint discovery with temporal prediction and object dynamics modeling \cite{Minderer2019UnsupervisedObjectStructureDynamics}, though in that formulation representation learning and dynamics modeling were deliberately decoupled to preserve spatial structure, limiting the extent to which temporal objectives shape the learned representation.
 
The application of unsupervised keypoints to action recognition and human motion analysis remains relatively underexplored compared to supervised pose estimation. A systematic comparison of unsupervised and supervised keypoints for fall detection, and in particular an analysis of the specific conditions under which each approach has an advantage, has not been addressed in prior work. This gap motivates the controlled experimental analysis presented in this paper.
 
\section{Methods}

Our proposed fall detection framework consists of four main stages: person segmentation, keypoint detection, keypoint prediction, and classification. The overall goal is to transform video frames into compact motion-based representations and then classify each sequence as fall or non-fall. 

\subsection{Pipeline Overview}

\begin{figure}
\centering
\includegraphics[width=\textwidth]{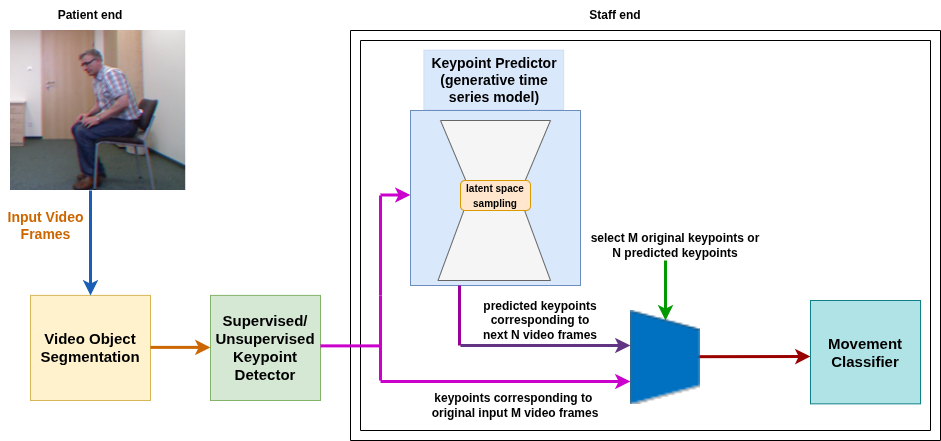}
\caption{The modular design of the proposed pipeline}\label{fig_2}
\end{figure}

The full pipeline takes a video sequence as input and processes it in several stages. First, person-centric regions are extracted from every video frame using a segmentation module. Next, each segmented frame is converted into a sequence of motion representations using either an unsupervised keypoint detector or a supervised pose detector. These keypoint trajectories are then passed to a temporal prediction model, which forecasts future motion over short horizons. The system only detects and transmits keypoints for the first $M$ frames, and a predictor is used to generate keypoints for the subsequent $N$
frames. Finally, a temporal classifier operates on the observed and/or predicted motion representation to determine whether the sequence corresponds to a fall or a non-fall event. This modular design enables a direct comparison between supervised and unsupervised keypoint front ends while keeping the downstream predictive and classification framework consistent. The design of the pipeline is shown in Figure \ref{fig_2}.
A key motivation for this design is that it replaces continuous raw-video transmission with compact motion summaries. As a result, the pipeline is naturally aligned with privacy-preserving and bandwidth-aware remote monitoring.

\subsection{YOLOv8 Segmentation}

To focus the pipeline on the human subject and reduce the influence of background clutter, a YOLOv8-m segmentation module was applied to the video frames. For each frame, the segmentation model isolates the person region from the surrounding scene \cite{ultralytics_yolov8_docs}. The segmented human region is then used as input to the keypoint detection module for both the supervised and unsupervised cases. This step is particularly important in fall detection because irrelevant background structures, furniture, and illumination changes may otherwise introduce noise into the learned motion representation. 
By constraining the analysis to the person region, the system emphasizes body configuration and motion dynamics that are most relevant to fall recognition.

\subsection{Keypoint Detector}
\subsubsection{Unsupervised Keypoint Detector}
\label{sec:3.3.1}

For unsupervised keypoint extraction, we adopt the keypoint detector architecture of \cite{Minderer2019UnsupervisedObjectStructureDynamics}. Their framework learns keypoints that capture the spatial structure of the object directly from video without manual pose annotations by using a keypoint-based representational bottleneck. Each video frame is passed through a convolutional detector that produces $K$ feature maps, one for each keypoint, where $K$ is the number of keypoints. These feature maps are normalized and converted into spatial coordinates $(x,y)$ through expectation over the heatmaps, yielding a compact set of learned keypoints for every frame. The reconstruction network then uses the detected keypoints together with a reference frame to reconstruct the image, forcing the model to encode the scene structure through keypoints rather than through an unrestricted latent representation. Along with the coordinates, the detector also returns the mean intensity $\mu_k$ of each keypoint feature map. Therefore, the keypoint detector returns a sequence $\mathbf{kp}_{t,k} = \bigl(x_{t,k},\, y_{t,k},\, \mu_{t,k}\bigr)$ for time index $t \in \{1,\dots,T\}$ and keypoint index $k \in \{1,\dots,K\}$, where $(x_{t,k}, y_{t,k})$ denote the detected coordinates and $\mu_{t,k}$ denotes the corresponding keypoint confidence or activation score.

To train the keypoint detector, three losses are used: image reconstruction loss $\mathcal{L}_{\mathrm{image}}$, temporal separation loss $\mathcal{L}_{\mathrm{sep}}$ and sparsity loss $\mathcal{L}_{\mathrm{sparse}}$.

\begin{equation}
    \mathcal{L}_{\mathrm{image}} = \sum_t \left\| \mathbf{v} - \hat{\mathbf{v}} \right\|_2^2
\end{equation}
where $\mathbf{v}$ is the true and $\hat{\mathbf{v}}$ is the reconstructed video frame.

\begin{equation}
\mathcal{L}_{\mathrm{sep}} = \sum_k \sum_{k'} \exp\!\left(-\frac{\frac{1}{T}\sum_t
\left\|
\left(\mathbf{kp}_{t,k} - \langle \mathbf{kp}_k \rangle\right)
-
\left(\mathbf{kp}_{t,k'} - \langle \mathbf{kp}_{k'} \rangle\right)
\right\|_2^2}{2\sigma_{\mathrm{sep}}^2}\right)
\label{eq:lsep}
\end{equation}
where $k$ and $k'$ are two different keypoint trajectories, $\langle \mathbf{kp}_{k} \rangle$ and $\langle \mathbf{kp}_{k'} \rangle$ are temporal mean of keypoint trajectory of $k$ and $k'$ respectively, and $\sigma_{\mathrm{sep}}$ is a Gaussian radius.

\begin{equation}
    \mathcal{L}_{\mathrm{sparse}} = \sum_k |\mu_k|
\end{equation}

The temporal separation loss and sparsity loss improve object
tracking performance and stability of the keypoints. The temporal separation loss encourages different keypoints to follow decorrelated trajectories over time, while the sparsity loss encourages parsimonious activation of keypoints. These inductive biases help the detector learn stable motion-sensitive landmarks that can represent object structure and movement effectively.

Because some keypoints may be weakly detected or missing in individual frames, we first apply temporal linear interpolation to the spatial coordinates before classification. A keypoint is considered valid whenever $\mu_{t,k} > \tau$, where $\tau$ is a confidence threshold (set to $0.05$ in our implementation). The value of $\tau$ was chosen as a conservative lower bound to identify only near-zero confidence detections as unreliable while preserving most keypoint observations.
Let $m_{t,k}$ denote the validity of the keypoint. Therefore,

\begin{equation}
m_{t,k} =
\begin{cases}
1, & \mu_{t,k} > \tau,\\
0, & \text{otherwise},
\end{cases}
\label{eq:valid_mask}
\end{equation}

\noindent For each temporal trajectory $x_{:,k}$ and $y_{:,k}$, let
\[
t^-_{t,k} = \max \{ s \le t \;:\; m_{s,k}=1\},
\qquad
t^+_{t,k} = \min \{ s \ge t \;:\; m_{s,k}=1\},
\]
denote the nearest previous and next valid time indices, respectively. If both neighbors exist, the missing coordinate is filled by linear interpolation. For the $x$-coordinate,
\begin{equation}
\tilde{x}_{t,k} =
x_{t^-_{t,k},k} +
\frac{t - t^-_{t,k}}{t^+_{t,k} - t^-_{t,k}}
\left(
x_{t^+_{t,k},k} - x_{t^-_{t,k},k}
\right),
\label{eq:x_interp}
\end{equation}
and similarly for the $y$-coordinate,
\begin{equation}
\tilde{y}_{t,k} =
y_{t^-_{t,k},k} +
\frac{t - t^-_{t,k}}{t^+_{t,k} - t^-_{t,k}}
\left(
y_{t^+_{t,k},k} - y_{t^-_{t,k},k}
\right).
\label{eq:y_interp}
\end{equation}

\noindent If only one valid neighbor exists, its value is propagated to fill the missing point. If the original keypoint is already valid, its detected coordinate is kept unchanged. Thus, the interpolated coordinates can be written compactly as
\begin{equation}
\hat{x}_{t,k} =
\begin{cases}
x_{t,k}, & m_{t,k}=1,\\
\tilde{x}_{t,k}, & m_{t,k}=0,
\end{cases}
\qquad
\hat{y}_{t,k} =
\begin{cases}
y_{t,k}, & m_{t,k}=1,\\
\tilde{y}_{t,k}, & m_{t,k}=0.
\end{cases}
\label{eq:interp_final}
\end{equation}

\noindent In our implementation, the confidence value $\mu_{t,k}$ is retained after interpolation so that the classifier still has access to the original detector confidence information. After interpolation, the keypoint trajectories are normalized to reduce variability due to absolute position and scale. 

Let $\hat{x}_{t,k}$ and $\hat{y}_{t,k}$ denote the interpolated coordinates. When centering over both time and keypoints, the spatial means are
\begin{equation}
\bar{x} = \frac{1}{TK}\sum_{t=1}^{T}\sum_{k=1}^{K}\hat{x}_{t,k},
\qquad
\bar{y} = \frac{1}{TK}\sum_{t=1}^{T}\sum_{k=1}^{K}\hat{y}_{t,k}.
\label{eq:global_mean}
\end{equation}
The centered coordinates are then
\begin{equation}
x'_{t,k} = \hat{x}_{t,k} - \bar{x}
\qquad
y'_{t,k} = \hat{y}_{t,k} - \bar{y}.
\label{eq:centered_coords}
\end{equation}

To normalize the spatial scale, we compute the root-mean-square magnitude across the full video:
\begin{equation}
s^x = \sqrt{\frac{1}{TK}\sum_{t=1}^{T}\sum_{k=1}^{K}(x'_{t,k})^2 + \varepsilon},
\qquad
s^y = \sqrt{\frac{1}{TK}\sum_{t=1}^{T}\sum_{k=1}^{K}(y'_{t,k})^2 + \varepsilon},
\label{eq:rms_scale}
\end{equation}
where $\varepsilon$ is a small constant for numerical stability.

The final normalized coordinates are therefore
\begin{equation}
x^{\mathrm{norm}}_{t,k} = \frac{x'_{t,k}}{s^x},
\qquad
y^{\mathrm{norm}}_{t,k} = \frac{y'_{t,k}}{s^y}.
\label{eq:normalized_coords}
\end{equation}

The normalized keypoint representation passed to the classifier is
\begin{equation}
\mathbf{k}^{\mathrm{norm}}_{t,k} =
\bigl(
x^{\mathrm{norm}}_{t,k},\;
y^{\mathrm{norm}}_{t,k},\;
\mu_{t,k}
\bigr).
\label{eq:normalized_kp}
\end{equation}

This preprocessing serves two purposes. First, temporal interpolation reduces the effect of intermittent missed detections by enforcing continuity in each keypoint trajectory. Second, normalization removes global translation and scale differences across videos, encouraging the classifier to focus on relative motion patterns and posture evolution rather than absolute image position. 

In our pipeline, the unsupervised keypoint detector is used to obtain a privacy-preserving and compact motion representation of the person. Since these keypoints are learned from the video data itself, they are not constrained to match anatomical joints exactly; rather, they act as motion-relevant landmarks that capture the structure and temporal evolution of the human body during activities and falls.

\subsubsection{Supervised keypoint Detector}
\label{sec:3.3.2}

We evaluate the supervised condition using OpenPose \cite{Cao2021OpenPose}, which serves as a representative and widely adopted supervised pose estimator and provides an established, anatomically interpretable reference point against which to assess the unsupervised representation. Unlike the unsupervised approach, OpenPose relies on pretrained human pose estimation models to detect 25 semantically meaningful body joints such as shoulders, elbows, hips, knees, and ankles. These supervised keypoints provide an explicit skeleton-based representation of human posture and movement. For each frame $t$ and joint $j$, the detector returns image coordinates, and a confidence score, denoted by $\mathbf{kp}_{t,j} = (x_{t,j}, y_{t,j}, \mu_{t,j})$. Before classification, since some joints may be missing or unreliable in individual frames, low-confidence detections are first removed using a confidence threshold of 0.15. A conservative threshold of 0.15 was adopted in this study which is high enough to exclude the most unreliable detections, but low enough to retain keypoints detected under occlusion with sufficient confidence \cite{Cao2021OpenPose}. Missing joints are then filled using linear interpolation over time as described in Section \ref{sec:3.3.1}. Since OpenPose is trained on individual frames with no temporal objective, a 5-frame simple moving average is applied to the supervised keypoint trajectories following interpolation to reduce temporal jitter \cite{martini2025denoising, lamaakal2025tiny}. Then, to reduce sensitivity to absolute image position, the keypoints are centered with respect to a reference joint, chosen as the MidHip when available. The centered keypoints are then scale-normalized using a body reference distance, chosen as the Neck-to-MidHip distance when available, and otherwise from the distance between the left and right shoulders \cite{tumidhip, zolfaghari2024sensor, de2023towards}. Then, the confidence score of each frame and joint is dropped and frame-to-frame coordinate difference is calculated for every joint \cite{li2018co}. Finally, both normalized joint coordinates and frame-to-frame displacements are concatenated to form the supervised pose representation used for classification. This preprocessing improves temporal continuity and makes the supervised representation more robust to missing detections, translation, and scale variation.

The inclusion of OpenPose allows us to compare a supervised pose-based representation with the unsupervised learned landmark representation. In the proposed framework, OpenPose keypoints serve as a structured baseline with anatomically interpretable joints, while the unsupervised detector provides a data-driven alternative that may better adapt to motion cues in domain-shifted video settings.

\subsection{Keypoint Predictor}

For future keypoint prediction, we use the stochastic dynamics model introduced in \cite{Minderer2019UnsupervisedObjectStructureDynamics}. After encoding each frame into a low-dimensional keypoint representation, the temporal evolution of the keypoints is modeled in coordinate space using a variational recurrent neural network (VRNN) \cite{Chung2015VRNN}. We use a VRNN rather than a standard recurrent neural network (RNN) because the temporal evolution of human motion is inherently uncertain and multimodal. In fall detection, similar observed motion prefixes can lead to different plausible future trajectories depending on subject behavior, speed, camera viewpoint, occlusion, and noise in the detected keypoints. A conventional RNN models the sequence dynamics deterministically through a single hidden-state trajectory, which can limit its ability to represent uncertainty and variability in future motion \cite{haque2026towards}. By contrast, the VRNN introduces a latent stochastic variable at each timestep, allowing the model to capture a distribution over possible temporal evolutions rather than a single deterministic prediction.

This property is especially important in our setting because our goal is not only to model sequence dynamics, but also to regularize the learned keypoint representation so that it captures motion patterns that are predictive of future behavior. VRNN jointly optimizes the recurrent state, latent variables, and reconstruction objective for end-to-end forecasting, thereby enabling temporal dynamics and stochastic variation to be learned in a coordinated and task-relevant manner \cite{haque2026towards, haque2025inference}. By encouraging the latent state to encode both the current observation and the uncertainty associated with future motion, VRNN provides a useful representation for fall prediction, which is particularly useful for distinguishing falls from non-fall activities that may appear similar in individual frames but differ in their temporal progression. 

In this model, a latent stochastic variable $\mathbf{z}$ captures uncertainty in future motion, while the recurrent hidden state $\mathbf{h}_{t-1}$ summarizes the temporal context from previous timesteps. The prior distribution over the latent variable $p(\mathbf{z}_t \mid \mathbf{kp}_{<t}, \mathbf{z}_{<t})$ is conditioned on the past hidden state, and the posterior $q(\mathbf{z}_t \mid \mathbf{kp}_{\leq t}, \mathbf{z}_{<t})$ combines the hidden state with the current detected keypoints $\mathbf{kp}_t = \bigl(x_{t},\, y_{t},\, \mu_{t}\bigr)$ as shown in equations \ref{eq:prior} and \ref{eq:post}.

\begin{equation}
    p(\mathbf{z}_t \mid \mathbf{kp}_{<t}, \mathbf{z}_{<t}) = \varphi^{\mathrm{prior}}(\mathbf{h}_{t-1})
\label{eq:prior}
\end{equation}

\begin{equation}
   q(\mathbf{z}_t \mid \mathbf{kp}_{\leq t}, \mathbf{z}_{<t}) = \varphi^{\mathrm{enc}}(\mathbf{h}_{t-1}, \mathbf{kp}_t) 
\label{eq:post}
\end{equation}

\noindent where $\varphi^{\mathrm{prior}}$ and $\varphi^{\mathrm{enc}}$ denote the neural networks for prior distribution and encoder respectively. 

The distribution of future keypoints $p(\mathbf{kp}_t \mid \mathbf{z}_{\leq t}, \mathbf{kp}_{<t})$ are then predicted by decoding from the latent state of current timestep $\mathbf{z}_t$ and recurrent context which is updated at each timestep as shown in equations \ref{eq:dec} and \ref{eq:rnn}.

\begin{equation}
   p(\mathbf{kp}_t \mid \mathbf{z}_{\leq t}, \mathbf{kp}_{<t}) = \varphi^{\mathrm{dec}}(\mathbf{z}_t, \mathbf{h}_{t-1})
\label{eq:dec}
\end{equation}

\begin{equation}
    h_t = \varphi^{\mathrm{RNN}}(\mathbf{kp}_t, \mathbf{z}_t, \mathbf{h}_{t-1})
\label{eq:rnn}
\end{equation}

\noindent where $\varphi^{\mathrm{dec}}$ denote the neural network for decoder and $\varphi^{\mathrm{RNN}}$ denote the recurrent update function, implemented using an RNN unit.

The standard VRNN is trained to encode the detected keypoints by maximizing the evidence lower bound (ELBO), which is composed of a reconstruction loss term and a Kullback-Leibler (KL) term between the Gaussian prior and posterior distributions:
\begin{equation}
\mathcal{L}_{\mathrm{VRNN}}
=
-\sum_{t=1}^{T_0}
\mathbb{E}\!\left[
\log p(\mathbf{kp}_t \mid \mathbf{z}_{\leq t}, \mathbf{kp}_{<t})
-
\beta \, \mathrm{KL}\!\left(
\mathcal{N}^{\mathrm{enc}}_t \,\|\, \mathcal{N}^{\mathrm{prior}}_t
\right)
\right]
\label{eq:lvrnn}
\end{equation}
where $\mathcal{N}^{\mathrm{enc}}_t = \mathcal{N}(\mathbf{z}_t \mid \varphi^{\mathrm{enc}}(\mathbf{h}_{t-1}, \mathbf{kp}_t))$ and $\mathcal{N}^{\mathrm{prior}}_t = \mathcal{N}(\mathbf{z}_t \mid \varphi^{\mathrm{prior}}(\mathbf{h}_{t-1}))$.

The KL regularization in the VRNN enforces consistency between the prior belief inferred from past motion and the posterior belief inferred after observing the current keypoints. This makes the temporal model learn not only how to reconstruct observed motion, but also to anticipate plausible future motion in a probabilistic manner. In contrast, a standard RNN trained only with deterministic prediction loss may overfit to average trajectories or frame-level artifacts, especially when the motion pattern is ambiguous. Therefore, the VRNN is better aligned with our objective of learning dynamically consistent, uncertainty-aware, and task-relevant motion representations for fall detection.

To encourage learning of long-term dependencies, a reconstruction loss is added without the KL term for multiple future timesteps:
\begin{equation}
\mathcal{L}_{\mathrm{future}}
=
-\sum_{t=T_0+1}^{T_0+\Delta T}
\mathbb{E}\!\left[
\log p(\mathbf{kp}_t \mid \mathbf{z}_{\leq t}, \mathbf{kp}_{\leq T_0})
\right]
\label{eq:lfuture}
\end{equation}

\noindent where $T_0$ and $\Delta T$ are the number of observed timesteps and the number of predicted timesteps respectively. 

This design enables prediction directly in keypoint coordinate space rather than in pixel space. The loss functions used to train the VRNN allow the model to capture both short-term consistency and longer-term motion uncertainty.

\subsection{Classifier Module}

The final classification stage uses a Long Short Term Memory (LSTM)-based classifier \cite{hochreiter1997long}. The LSTM receives the temporal sequence of keypoint features, either directly detected or predicted using the VRNN, and learns the sequential motion patterns associated with fall and non-fall activities. Because LSTMs are designed to model temporal dependencies, they are well suited for distinguishing gradual daily activities from the rapid posture transitions and motion irregularities that characterize falls. The classifier operates on the keypoint sequence rather than raw pixels, which significantly reduces input dimensionality and emphasizes motion structure. This also improves privacy by avoiding the need to process detailed appearance information at the decision stage. The LSTM outputs the final class label indicating whether the input sequence corresponds to a fall or a non-fall event.

Given an input keypoint sequence $\{\mathbf{kp}_{1,k}, \mathbf{kp}_{2,k}, \dots, \mathbf{kp}_{T,k}\}$, the LSTM produces hidden states at each timestep
and the final hidden state $\mathbf{h}_T$ is used as the sequence representation for classification. The classifier outputs class probabilities $\hat{\mathbf{y}} = \mathrm{softmax}\!\left(\mathbf{W}_{\mathrm{cls}} \mathbf{h}_T + \mathbf{b}_{\mathrm{cls}}\right)$
where $\hat{\mathbf{y}} \in \mathbb{R}^{2}$, i.e., each element of $\hat{\mathbf{y}}$ represents the predicted probability of one class. The model is trained using the cross-entropy loss
\begin{equation}
\mathcal{L}_{\mathrm{cls}} = - \sum_{c=1}^{2} y_c \log \hat{y}_c
\label{eq:lcls}
\end{equation}
where $c$ indexes the class label for the fall and non-fall. Here, $\mathbf{y}$ is the one-hot ground-truth label and $\hat{\mathbf{y}}$ is the predicted class-probability vector.

\subsection{Our Proposed Pipeline}
\label{sec:3.6}

The raw keypoint detector (both supervised and unsupervised) outputs a sequence for each video $\mathbf{kp}_{t,k} = \bigl(x_{t,k},\, y_{t,k},\, \mu_{t,k}\bigr)$, for time index $t \in \{1,\dots,T\}$ and keypoint index $k \in \{1,\dots,K\}$ and the keypoint predictor receives the observed keypoint sequence from the first $M$ frames $
\{\mathbf{kp}_{t,k}\}_{t=1}^{M}$
and forecasts keypoints for the next $N$ frames
$\{\hat{\mathbf{kp}}_{t,k}\}_{t=M+1}^{M+N}$.
Thus, for a window of $M+N$ frames, the sequence becomes
\begin{equation}
\mathbf{kp}^{\mathrm{seq}}_{t,k} =
\begin{cases}
\mathbf{kp}_{t,k}, & 1 \leq t \leq M,\\[4pt]
\hat{\mathbf{kp}}_{t,k}, & M+1 \leq t \leq M+N,
\end{cases}
\label{eq:combined_kp_sequence}
\end{equation}
where $\hat{\mathbf{kp}}_{t,k}$ denotes the predicted keypoint at future timestep $t$. 
That is, the video is processed as consecutive non-overlapping blocks of length $M+N$, where the first $M$ frames in each block are observed and the following $N$ frames are predicted. The per-window observed and predicted keypoints are concatenated into a single sequence spanning the video, which is then preprocessed as described in Sections \ref{sec:3.3.1} and \ref{sec:3.3.2} before the downstream classification.

For the pipeline with supervised keypoint detector, the classifier receives normalized joint coordinates and their temporal differences as input of each video and classifies as fall or non-fall. Since OpenPose is used as a pretrained and fixed keypoint detector, only the keypoint predictor and the classifier are trained jointly.

On the other hand, for the pipeline with unsupervised keypoint detector, the LSTM classifier takes the keypoint sequence as represented in Eq. \ref{eq:normalized_kp} as input.

The total loss for training the pipeline with unsupervised keypoint detector is:

\begin{equation}
    \mathcal{L}_{\mathrm{total}} = \mathcal{L}_{\mathrm{image}} + \lambda_{sep}\mathcal{L}_{\mathrm{sep}} + \lambda_{sparse}\mathcal{L}_{\mathrm{sparse}} + \mathcal{L}_{\mathrm{VRNN}} + \mathcal{L}_{\mathrm{future}} + \lambda_{cls}\mathcal{L}_{\mathrm{cls}}
\label{eq:ltotal}
\end{equation}

\noindent where $ \lambda_{sep}$, $ \lambda_{sparse}$ and $ \lambda_{cls}$ are scale parameters for keypoint separation, sparsity and classification losses respectively.

The central goal of this design is not to recover the anatomically correct human pose in a conventional sense, but rather to learn a motion representation that is useful for fall detection. In many vision pipelines, performance is implicitly tied to how accurately the model localizes semantically meaningful body joints such as elbows, knees, and ankles. However, for fall detection in real-world healthcare settings, such anatomical correctness is not necessarily the right objective. In practical deployment, patient videos may contain occlusion, partial body visibility, or subject-specific appearance variation. Under these conditions, semantically grounded pose estimation can become unstable, and errors in joint localization may propagate to the downstream classifier. For fall detection, the more important requirement is not whether each detected point corresponds to a true anatomical landmark, but whether the representation preserves the temporal evolution of posture and motion that distinguishes falls from non-fall activities.

Accordingly, the objective in \eqref{eq:ltotal} is designed to encourage the emergence of keypoints that are \emph{task-aligned}. The image reconstruction loss $\mathcal{L}_{\mathrm{image}}$ ensures that the representation remains grounded in the visual content of the input and retains enough spatial information to reconstruct the observed frame. The separation and sparsity terms, $\mathcal{L}_{\mathrm{sep}}$ and $\mathcal{L}_{\mathrm{sparse}}$, discourage redundant or overly dense keypoint usage, thereby pushing the detector toward a compact set of motion-sensitive landmarks. On their own, however, these losses would still primarily favor frame-wise spatial structure.

The terms $\mathcal{L}_{\mathrm{VRNN}}$ and $\mathcal{L}_{\mathrm{future}}$ do not merely train a forecasting module; they act as temporal regularizers on the keypoint space itself. Because the latent dynamics model must explain the observed sequence and predict future motion, the learned keypoints are encouraged to encode features that evolve consistently through time. In other words, the representation is pressured to retain motion cues that are dynamically predictive.
This is particularly important for fall detection, since many hard cases involve activities such as sitting, bending, lying down, or transitional Activities of Daily Living (ADL) movements that may resemble falls in individual frames but differ substantially in their temporal progression.


These considerations motivate a design choice in which the keypoint detector, the temporal predictor, and the classifier are optimized together rather than in isolation. We note that Minderer et al. \cite{Minderer2019UnsupervisedObjectStructureDynamics} made the opposite choice, decoupling keypoint discovery from temporal prediction so that the representation is learned primarily from image-level reconstruction; that choice suits their goal of preserving general-purpose spatial structure. Our objective is different: we seek a representation shaped by the specific demands of fall detection, and coupling the objectives allows the temporal and classification losses to influence the keypoints themselves rather than only the modules downstream of them. This flexibility exists only because the unsupervised detector is learnable. A supervised pose detector such as OpenPose is used here as a fixed, pretrained module whose keypoints are determined by an anatomical-accuracy objective, before and independently of the fall-detection task; its front-end representation cannot be adapted to the task in the same way. The unsupervised approach therefore has a degree of design freedom at the detector stage that the supervised pipeline does not.

The classification loss $\mathcal{L}_{\mathrm{cls}}$ provides the final task-level supervision that completes this alignment. While the reconstruction and regularization losses keep the keypoints compact and visually grounded, and the VRNN terms encourage temporal consistency and predictiveness, the classification term explicitly rewards representations that separate falls from non-fall activities. Thus, the learned keypoints are shaped by three complementary pressures: spatial faithfulness, temporal consistency, and task discriminability. The resulting keypoints need not correspond one-to-one with human joints. Instead, they serve as compact motion anchors that summarize parts of the visual scene whose temporal behavior is most informative for distinguishing falls from non-fall activities. The whole fall detection framework using unsupervised keypoint detector is described in Algorithm \ref{algo:fall_pipeline} and the hyperparameters used for training the pipeline are given in Appendix Table A1.


\begin{algorithm}
\caption{Fall detection pipeline using unsupervised keypoint detector}
\label{algo:fall_pipeline}
\begin{algorithmic}[1]
\Require Video $V=\{I_t\}_{t=1}^{T}$, observed length $M$, predicted length $N$, confidence threshold $\tau$
\Ensure Fall / Non-fall label $c$
\State Extract raw keypoints from each frame:
\[
\mathbf{kp}_{t,k} = (x_{t,k}, y_{t,k}, \mu_{t,k}), \quad t=1,\dots,T,\; k=1,\dots,K
\]
\State Set window length $W \Leftarrow M + N$
\State Set number of windows $J \Leftarrow \lfloor T / W \rfloor$
\State Set assembled sequence length $T' \Leftarrow JW$
\State Initialize sequence $\mathcal{S} \Leftarrow \emptyset$
\For{$j = 0$ to $J-1$}
    \State Extract observed segment
    \[
    \mathcal{O}^{(j)} = \{\mathbf{kp}_{t,k}\}_{t=jW+1}^{jW+M}
    \]
    \State Predict future keypoints for next $N$ frames
    \[
    \mathcal{P}^{(j)} = \{\hat{\mathbf{kp}}_{t,k}\}_{t=jW+M+1}^{jW+W}
    \]
    \State Append observed and predicted keypoints to the sequence
    \[
    \mathcal{S} \Leftarrow \mathcal{S} \cup \mathcal{O}^{(j)} \cup \mathcal{P}^{(j)}
    \]
\EndFor
\For{each keypoint trajectory $k = 1,\dots,K$ in $\mathcal{S}$}
    \For{each timestep $t = 1,\dots,T'$}
        \If{$\mu_{t,k} \le \tau$}
            \State Mark keypoint $(x_{t,k}, y_{t,k})$ as invalid
        \EndIf
    \EndFor
    \State Linearly interpolate invalid $x$- and $y$-coordinates over time
\EndFor
\State Normalize the interpolated sequence over the full video to obtain
\[
\mathbf{kp}^{\mathrm{norm}}_{t,k} =
\bigl(
x^{\mathrm{norm}}_{t,k},\;
y^{\mathrm{norm}}_{t,k},\;
\mu_{t,k}\bigr),
\quad t = 1,\dots,T'
\]
\State Feed the normalized sequence $\mathbf{kp}^{\mathrm{norm}}$ to the LSTM classifier
\State Obtain predicted class probabilities $\hat{\mathbf{y}}$
\State $c \Leftarrow \arg\max \hat{\mathbf{y}}$
\State \Return $c$
\end{algorithmic}
\end{algorithm}

\section{Results}\label{sec2}

\subsection{Study Overview}

We evaluated the proposed fall-monitoring pipeline under two complementary experimental protocols. The first was an in-distribution baseline using random splitting, designed to establish general performance when no systematic difference exists between training and test conditions. The second was a set of out-of-distribution evaluations designed to assess generalization under conditions that differ from training in a controlled and meaningful way. Two out-of-distribution settings were examined: subject-wise generalization, where the model is tested on individuals not seen during training, and occlusion-based generalization, where the model is trained on videos with full body visibility and tested on videos where the body is partially occluded in most of the frames. These settings reflect two distinct deployment challenges. Subject-wise evaluation captures the requirement that a real-world system must operate on new patients whose appearance, motion patterns, and movement style were not represented during development. Occlusion-based evaluation captures the structural degradation that occurs when body parts are outside the camera frame or self-occluded, which is common in room-based monitoring where patient movement is unconstrained. Across all settings, our goal was to compare whether unsupervised keypoints offer practical advantages over supervised pose-based keypoints as motion features for temporal modeling and fall classification, and to characterize the specific conditions under which each approach has an advantage.

\subsection{Datasets}

We evaluated the proposed pipeline on two publicly available fall-analysis datasets: the UR Fall Detection dataset \cite{kwolek2014human} and the Human Fall Detection/GMDCSA-24 dataset \cite{alam2024gmdcsa}, which serve complementary roles in our experimental design. Both datasets capture indoor human activities and fall events from fixed-camera viewpoints, consistent with the camera placement expected in hospital rooms, assisted-living spaces, and home-monitoring environments.

The \textbf{Human Fall dataset} is used to evaluate subject-wise generalization. It contains 81 fall clips and 79 ADL clips performed by four subjects across three natural home environments, recorded under varying conditions including different activities and clothing across sessions. The videos range from 3 to 17 seconds in length. The availability of explicit subject labels makes this dataset well-suited for leave-one-subject-out evaluation, where generalization to new individuals is the primary question.

The \textbf{UR Fall dataset} is used to evaluate occlusion-based generalization. It contains 30 fall sequences and 40 ADL sequences recorded indoors, with videos ranging from 2 to 12 seconds. Unlike the Human Fall dataset, the UR Fall dataset does not provide explicit subject identity labels. However, the fall sequences are reported to involve five individuals \cite{kwolek2014human}; we recovered the assignment of videos to these identities by manual inspection. These five fall identities were used to construct identity-disjoint train and test splits for the occlusion experiment, as described in Section \ref{sec:4.4.3}.

\subsection{Data Augmentation}

To improve robustness to visual variability and increase the effective training set size, we applied image-based augmentation to each video. Augmented variants were generated by adjusting brightness, saturation, contrast, and sharpness, applying mild Gaussian blur and noise, performing gamma correction, horizontal flipping, and rotation. Due to the smaller size of the UR Fall dataset, 16 augmented variants were generated per original video, compared to 8 variants per video for the Human Fall dataset. The augmented videos were included only in training and validation sets in each experimental setup; test sets always consisted of original unaugmented videos to preserve the authenticity of the evaluation conditions.

\subsection{Experimental Protocol}

\subsubsection{Random Split (In-Distribution Baseline)}

For the in-distribution baseline, each dataset was split at the source-video level into training and test sets using an 80:20 ratio stratified by class label, so that the fall-to-ADL ratio of the full dataset was preserved in both partitions, with no control imposed over subject identity or visual conditions. This protocol reflects the standard evaluation practice in much of the fall detection literature and is included to establish a reference point for comparison with the controlled out-of-distribution experiments. The test set was held out throughout model development. The remaining 80\% training portion was divided into five folds for cross-validation. In each fold, four folds were used for training and one fold was used for validation. Since no identity constraint was applied in this setting, validation folds were stratified by class labels so that each fold contained approximately the same proportion of the total fall samples and the total ADL samples.

\subsubsection{Subject-Wise Generalization (Human Fall Dataset)}
\label{sec:4.4.2}

For the subject-wise out-of-distribution evaluation, we adopted a leave-one-subject-out protocol on the Human Fall dataset, where one group of videos of a subject was held out as the test set and the remaining three groups were used for model development. Each `group' denotes the collection of videos belonging to a single subject; holding out a group therefore corresponds to holding out one individual entirely. Within the development set, three-fold cross-validation was performed, with each fold using a different one of the three remaining groups as the validation set and the other two for training. This procedure was repeated for each of the four subject groups as the held-out test group, yielding four independent evaluations. Augmented videos of the non-held-out subjects were included in training and validation sets, while only original unaugmented videos of the held-out subject were used for testing. This resulted in approximately 1000 training and validation videos and around 40 test videos per experiment.

\subsubsection{Occlusion-Based Generalization (UR Fall Dataset)}
\label{sec:4.4.3}

For the occlusion-based out-of-distribution evaluation, we constructed a condition-controlled split on the UR Fall dataset designed to isolate body visibility as the experimental variable through manual video inspection. Training videos consisted of sequences where the subject's full body was predominantly visible throughout the video, including augmented variants of those videos. Test videos consisted of original unaugmented sequences where the body was partially visible for the majority of the video, due to self-occlusion or the subject moving partially outside the camera frame. To prevent identity-based confounds, the five manually identified fall identities were divided into disjoint training and test groups, with three identities used for training falls and the remaining two reserved for test falls. For ADL sequences, despite the absence of explicit identity labels, manual inspection confirmed that subjects appearing in ADL sequences across both train and test splits were recorded in different clothing, reducing the risk of appearance-based identity leakage. The training and validation set comprised 560 videos after augmentation and the test set comprised 18 videos.

For validation, five-fold cross-validation was performed on the training set using a video-level random split. Folds were stratified by both identity and class label for fall videos, and by class label only for ADL videos, to ensure balanced representation of all training fall identities across folds.

\subsubsection{Checkpoint Selection and Reporting}

Across all experiments and split types, the best model checkpoint was selected based on the highest validation accuracy achieved during training. The classification threshold was subsequently determined by identifying the value that maximized the F1 score on the same validation set. After each training run, the selected checkpoint was evaluated on the fixed held-out test set using the selected threshold. Reported metrics include accuracy, precision, recall, and F1 score, presented as mean $\pm$ standard deviation across folds. Since fall detection involves an inherent class imbalance between fall and ADL sequences, F1 score is used as the primary evaluation metric, as it provides a more reliable measure of performance than accuracy under imbalanced conditions.

\subsection{Comparison of Unsupervised and Supervised keypoints on Fall Detection}

 In order to compare the performance of supervised and unsupervised keypoints on fall detection, we conducted experiments on UR Fall dataset and Human Fall dataset using pipelines with both supervised and unsupervised keypoint detectors. The first set of experiments was performed without the keypoint predictor which means keypoints were extracted from every frame using the keypoint detector and sent to the classifier after preprocessing. The second set of experiments included the keypoint predictor and in this case to emulate bandwidth-constrained operation half of the total number of frames were predicted instead of transmitting keypoints from every frame following the method as described in Section \ref{sec:3.6}. 



\subsubsection{Random Split}
\begin{table}[h]
\caption{Comparison of classification performance (mean ± standard deviation across 5 validation folds for a representative seed) using pipelines with supervised and unsupervised keypoints on random split of UR Fall dataset and Human Fall dataset. The upper panel and the lower panel demonstrate results without and with prediction respectively.}
\label{tab:random}
\centering
\setlength{\tabcolsep}{12pt}
\scriptsize
\begin{tabular}{@{}lcccccccc@{}}
\toprule
& \multicolumn{2}{c}{Accuracy (\%)} 
& \multicolumn{2}{c}{Precision (\%)} 
& \multicolumn{2}{c}{Recall (\%)} 
& \multicolumn{2}{c}{F1-Score (\%)} \\
\cmidrule(lr){2-3} \cmidrule(lr){4-5} \cmidrule(lr){6-7} \cmidrule(lr){8-9}
&&&& \textbf{Without} & \textbf{prediction} && \\
\midrule
Dataset 
& Unsup.\footnotemark[1] & Sup.\footnotemark[1]
& Unsup. & Sup. 
& Unsup. & Sup. 
& Unsup. & Sup. \\
\midrule
UR Fall      & \textbf{95.7}±5.7 & 82.9±5.7 & \textbf{93.8}±7.6 & 75±5.5 & \textbf{96.7}±6.7 & 90±8.2 & \textbf{95.1}±6.6 & 81.8±6.3 \\
Human Fall   & 84.4±2 & \textbf{95}±1.5 & 88.1±3.6 & \textbf{92}±2.5 & 80±7.3 & \textbf{98.5}±2.5 & 83.5±2.9 & \textbf{95.2}±1.5 \\
\bottomrule
&&&& \textbf{With} & \textbf{prediction} && \\
\midrule
Dataset
& Unsup.\footnotemark[1] & Sup.\footnotemark[1]
& Unsup. & Sup. 
& Unsup. & Sup. 
& Unsup. & Sup. \\
\midrule
UR Fall      & \textbf{91.4}±2.9 & 84.3±5.3 & \textbf{93.8}±7.6 & 77.4±5& \textbf{86.7}±6.7 & 90±13.3  & \textbf{89.7}±3.2 & 82.7±7 \\
Human Fall   & 87.5±2 & \textbf{91.9}±3.2 & 87.3±5 & \textbf{89.4}±2.4 & 88.7±9.2 & \textbf{95}±4.7 & 87.5±2.7 & \textbf{92.1}±3.2 \\
\bottomrule
\end{tabular}
\footnotetext[1]{Sup. refers to supervised and Unsup. refers to unsupervised.}
\end{table}

\noindent Under random splitting, the test set of UR Fall dataset consists of 8 ADL videos and 6 Fall videos, and the test set of Human Fall consists of 16 ADL videos and 16 Fall videos. To assess whether observed performance differences were consistent across training initializations rather than attributable to specific weight configurations, experiments were repeated with three random seeds for both datasets under both prediction settings. Table~\ref{tab:random} reports the full classification performance (accuracy, precision, recall, and F1) for a representative seed, with values given as mean ± standard deviation across validation folds, for both datasets, both prediction settings, and both the pipelines using unsupervised and supervised keypoints. Tables \ref{tab:f1_pairs_ur_rand} and \ref{tab:f1_pairs_hum_rand} report, for each of the three seeds, the mean F1 score across validation folds, giving six paired unsupervised-supervised comparisons per dataset (three seeds × two prediction settings).


These results indicate that under standard in-distribution evaluation, the relative behavior of the two representations varies by dataset and does not consistently favor either approach: on UR Fall dataset the unsupervised pipeline scores higher across seeds, whereas on Human Fall dataset the supervised pipeline scores higher. We do not attach a significance test to this baseline, because the random split imposes no controlled difference between training and test conditions and therefore does not correspond to a well-defined effect of interest. Since random splitting imposes no control over visual conditions, subject identity, or occlusion level, it cannot reveal the specific conditions under which each representation has an advantage.
These results therefore serve as an upper-bound reference rather than a meaningful discriminator between the two approaches. To characterize when each pipeline has a genuine advantage, we turn to out-of-distribution evaluations in which training and test conditions differ in a controlled and meaningful way: first by holding out unseen subjects, and then by introducing occlusion as a structured visual degradation. These controlled settings are more representative of real deployment conditions and, as we show, reveal clear and interpretable differences between the two keypoint representations.

\begin{table}[h]
\caption{Paired mean F1 scores across validation folds for three random seeds on the UR Fall dataset under the random split. The first and second value of the pair refer to the mean F1 score obtained from using the pipeline with unsupervised keypoint detector and supervised keypoint detector respectively.}
\label{tab:f1_pairs_ur_rand}
\centering
\begin{tabular*}{\textwidth}{@{\extracolsep{\fill}}lccc@{}}
\toprule
Prediction type
& \textbf{Seed\#1}
& \textbf{Seed\#2} 
& \textbf{Seed\#3} \\
\midrule
Without      & (\textbf{95.1}, 81.8) & (\textbf{98.5}, 85.3) & (\textbf{91.5}, 82.2) \\
With      & (\textbf{89.7}, 82.7) & (\textbf{88.6}, 80.3) & (\textbf{94.5}, 82.8) \\
\bottomrule
\end{tabular*}
\end{table}

\begin{table}[h]
\caption{Paired mean F1 scores across validation folds for three random seeds on the Human Fall dataset for random split. The first and second value of the pair refer to the F1 score obtained from using the pipeline with unsupervised keypoint detector and supervised keypoint detector respectively.}
\label{tab:f1_pairs_hum_rand}
\centering
\begin{tabular*}{\textwidth}{@{\extracolsep{\fill}}lccc@{}}
\toprule
Prediction type
& \textbf{Seed\#1}
& \textbf{Seed\#2} 
& \textbf{Seed\#3} \\
\midrule
Without      & (83.5, \textbf{95.2}) & (84.4, \textbf{93.5}) & (83, \textbf{91.5}) \\
With      & (87.5, \textbf{92.1}) & (84.6, \textbf{92.8}) & (85.9, \textbf{92.9}) \\
\bottomrule
\end{tabular*}
\end{table}

\subsubsection{Subject-Wise Generalization}

\begin{table*}[t]
\centering
\caption{Group-wise summary of ADL and fall activities in the Human Fall dataset.}
\label{tab:humanfall_subject_summary}
\begin{tabular}{p{1.6cm} p{5.2cm} p{5.2cm}}
\toprule
\textbf{Subject} & \textbf{ADL summary} & \textbf{Fall summary} \\
\midrule
Group 1 &
Mainly simpler daily activities centered on \textit{sitting, reading, walking, and sleeping} with truncated body poses for longer time. &
Contains all three fall directions, but is dominated by \textit{sideways falls}, with fewer forward and backward falls. Some clips include short pre-fall context such as sitting, standing, or walking. \\
\midrule
Group 2 &
Most behaviorally diverse ADL set, including \textit{drinking, eating, exercising, reading, sitting, sleeping, standing, walking, picking up and writing} with bending poses for a short time. &
Most diverse fall set, containing \textit{backward, forward, and sideways falls}, often preceded by contextual actions such as walking, drinking, eating, reading, sitting, standing, and writing. \\
\midrule
Group 3 &
Varied ADL set with strong emphasis on \textit{sitting}, along with drinking, eating, exercise, reading, sleeping, standing, walking, and writing. &
Dominated by \textit{backward falls}, with fewer forward and sideways falls. Includes falls from standing, sitting, and bed-related contexts, as well as some slow falls. \\
\midrule
Group 4 &
ADL set is centered on \textit{sitting}, together with drinking, eating, exercise, reading, sleeping, standing, walking, working on laptop and writing. &
Dominated by \textit{sideways} and \textit{forward falls}, with relatively few backward falls. Many clips involve sitting-on-bed to fall, bed-to-floor transitions, walking-to-fall, and chair-to-fall contexts with partial body visibility due to camera angle. \\
\bottomrule
\end{tabular}
\end{table*}

\begin{table}[h]
\caption{Comparison of classification performance (mean ± standard deviation across 3 validation folds for a representative seed) using pipelines with supervised and unsupervised keypoints on the Human Fall dataset for subject-wise generalization. The upper panel and the lower panel demonstrate results for without and with prediction respectively.}
\label{tab:humanfall_results}
\centering
\setlength{\tabcolsep}{12pt}
\scriptsize
\begin{tabular}{@{}lcccccccc@{}}
\toprule
& \multicolumn{2}{c}{Accuracy (\%)} 
& \multicolumn{2}{c}{Precision (\%)} 
& \multicolumn{2}{c}{Recall (\%)} 
& \multicolumn{2}{c}{F1-Score (\%)} \\
\cmidrule(lr){2-3} \cmidrule(lr){4-5} \cmidrule(lr){6-7} \cmidrule(lr){8-9}
&&&& \textbf{Without} & \textbf{prediction} && \\
\midrule
Test Group 
& Unsup.\footnotemark[1] & Sup.\footnotemark[1]
& Unsup. & Sup. 
& Unsup. & Sup. 
& Unsup. & Sup. \\
\midrule
Group 1      & \textbf{86.5}±6.5 & 84.4±4.4 & \textbf{83.7}±9.4 & 79.4±5.6 & 91.7±3.6 & \textbf{93.8}±0 & \textbf{87.3}±5.6 & 85.9±3.4 \\
Group 2      & 86.8±10.7 & \textbf{95.1}±2.6 & 82.3±12.4 & \textbf{91.7}±4.1 & 97.3±2.3 & \textbf{100}±0.0 & 88.9±8.2 & \textbf{95.6}±2.3 \\
Group 3      & 83.7±4 & \textbf{97.7}±1.9 & 86.2±3.1 & \textbf{95.6}±3.6 & 79.4±7.3 & \textbf{100}±0 & 82.5±4.9 & \textbf{97.7}±1.9 \\
Group 4      & \textbf{81.1}±4.7 & 75.7±5.8 & \textbf{75}±1.1 & 68.5±5.7 & \textbf{88.2}±15.6 & \textbf{88.2}±4.8 & \textbf{80.6}±6.9 & 77±5.1 \\
\bottomrule
&&&& \textbf{With} & \textbf{prediction} && \\
\midrule
Test Group 
& Unsup.\footnotemark[1] & Sup.\footnotemark[1]
& Unsup. & Sup. 
& Unsup. & Sup. 
& Unsup. & Sup. \\
\midrule
Group 1      & 77.1±2.9 & \textbf{87.5}±5.1 &72.7±3.2 & \textbf{83.4}±3.9 & 87.5±8.8 & \textbf{93.8}±8.8 & 79.1±3.4 & \textbf{88.1}±5.3 \\
Group 2      & 84±8 & \textbf{89.6}±0 & 78.2±7.8 & \textbf{84.1}±1.1 & 97.3±3.8 & \textbf{98.7}±1.9 & 86.6±6.3 & \textbf{90.8}±0.2 \\
Group 3      & 83.7±1.9 & \textbf{93}±1.9 & 85.2±2.7 & \textbf{90.1}±4.6 & 81±6.7 & \textbf{96.8}±4.5 & 82.8±2.7 & \textbf{93.1}±1.7 \\
Group 4      & 75.7±0 & \textbf{82.9}±3.4 & \textbf{77.8}±3.9 & 75.9±3.5 & 66.7±5.5 & \textbf{92.2}±2.8 & 71.5±1.8 & \textbf{83.2}±3.2 \\
\bottomrule
\end{tabular}
\footnotetext[1]{Sup. refers to supervised and Unsup. refers to unsupervised.}
\end{table}

Table \ref{tab:humanfall_subject_summary} contains the group-wise summary of ADL and Fall activities of Human Fall dataset. The experiments were repeated with three different random seeds under the same experimental settings on Human Fall dataset for subject-wise generalization. The full classification performance for four groups, for a representative seed under both prediction settings and the 24 paired mean F1 scores across validation folds (three seeds × two prediction settings × four groups) are shown in Table \ref{tab:humanfall_results} and Table \ref{tab:f1_pairs} respectively.

Although the experiment is designed to assess generalization to unseen subjects, the leave-one-subject-out protocol does not vary subject identity in isolation: each held-out subject also differs from the others in visual characteristics such as body framing, camera angle, degree of body visibility, and the variety of activities performed (Table \ref{tab:humanfall_subject_summary}). Subject identity and visibility condition therefore co-vary, and differences observed across held-out groups cannot be attributed to identity alone.




Without prediction, the supervised pipeline is stronger in Group 2, which consists of 23 ADL videos and 25 Fall videos and Group 3, which consists of 22 ADL videos and 21 Fall videos; as it is favored in all three seeds, by medians of 6.7 and 8.0 F1 points respectively. Whereas in Group 4, which consists of 20 ADL videos and 17 Fall videos, the unsupervised pipeline is ahead in all three seeds by median of 3.6 F1 points and in Group 1, which consists of 16 ADL videos and 16 Fall videos, the two are effectively indistinguishable. These groups differ in their dominant visual characteristics: Groups 2 and 3 are dominated by well-framed sequences in which anatomical landmarks are generally detectable, whereas Groups 1 and 4 are dominated by truncated body poses and partial body visibility arising from camera angle (Table \ref{tab:humanfall_subject_summary}). The relative performance of the two representations is thus associated with the prevalence of partial visibility within each group. Additionally, Groups 2 and 3 contain highly varied activities and abrupt posture transitions, which appear to pose a particular difficulty for the unsupervised representation: complex or rapidly changing non-fall motion is not always separated sharply from fall motion, which can produce additional false positives.

With prediction, the supervised pipeline is favored in eleven of the twelve group-seed comparisons, including in Group 4. Because each held-out evaluation trains the temporal model on the remaining groups, which include comparable partial-visibility sequences, the model can produce a more temporally coherent trajectory than the raw detections it replaces, so that forecasting partially compensates for unreliable supervised detection in Group 4.

\textbf{Two-sided Wilcoxon signed-rank test:} To assess whether the performance difference between the pipelines with unsupervised keypoint detector and supervised keypoint detector was statistically significant under subject-wise generalization, a paired two-sided Wilcoxon signed-rank test was applied \cite{wilcoxon1992individual} on the paired mean F1 scores (Table \ref{tab:f1_pairs}), using the difference between the scores for each pair. 

A Wilcoxon signed-rank test was used rather than a paired t-test because the small number of paired observations does not permit a reliable assessment of normality, and the test does not assume normally distributed paired differences, and a two-sided test was used because no directional hypothesis was specified for this comparison.
\begin{table}[h]
\caption{Paired mean F1 scores across validation folds for three random seeds on the Human Fall dataset for subject-wise generalization. The first and second value of the pair refer to the mean F1 score obtained from using the pipeline with unsupervised keypoint detector and supervised keypoint detector respectively. The upper panel and the lower panel demonstrate results for without and with prediction respectively.}
\label{tab:f1_pairs}
\centering
\begin{tabular*}{\textwidth}{@{\extracolsep{\fill}}lccc@{}}
\toprule
Test Group
& \textbf{Seed\#1}
& \textbf{Seed\#2} 
& \textbf{Seed\#3} \\
\midrule
\multicolumn{4}{c}{\textbf{Without prediction}} \\
\midrule
Group 1      & (\textbf{87.3}, 85.9) & (\textbf{89.6}, 85) & (86.2, \textbf{86.9}) \\
Group 2       & (88.9, \textbf{95.6}) & (88.5, \textbf{93.2}) & (87.1, \textbf{94.4}) \\
Group 3      & (82.5, \textbf{97.7}) & (86.5, \textbf{93.9}) & (87.5, \textbf{95.5}) \\
Group 4       & (\textbf{80.6}, 77) & (\textbf{77.3}, 75.6) & (\textbf{79.8}, 75.7) \\
\bottomrule
\multicolumn{4}{c}{\textbf{With prediction}} \\
\midrule
Test Group
& \textbf{Seed\#1} 
& \textbf{Seed\#2} 
& \textbf{Seed\#3} \\
\midrule
Group 1       & (79.1,\textbf{ 88.1}) & (83.6, \textbf{91.3}) & (84, \textbf{90.4}) \\
Group 2       & (86.6, \textbf{90.8})  & (\textbf{93.5}, 92.5) & (89.4, \textbf{91.5}) \\
Group 3       & (82.8, \textbf{93.1}) & (79.9, \textbf{91.7}) & (82.8, \textbf{91.7}) \\
Group 4       & (71.5, \textbf{83.2}) & (68.5, \textbf{82.4}) & (72.5, \textbf{82.1}) \\
\bottomrule
\end{tabular*}
\end{table}

The test result indicates a significant overall advantage for the supervised pipeline at the 5\% significance level (n = 24, W = 31.0, two-sided $p \approx  0.0003$, median 7.0 F1 points). However, we do not interpret it as evidence that supervised keypoints generalize better to new individuals, for two reasons. First, because identity and visibility are confounded in this design, the aggregate reflects the combined influence of both rather than subject generalization specifically. Second, the pooled effect is not homogeneous across conditions: when the two prediction settings are considered separately, the two representations are statistically indistinguishable without prediction (n = 12, median 2.7 F1 points, p = 0.15), whereas the supervised advantage is concentrated in the with-prediction setting (n = 12, median 9.0 F1 points, $p \approx 0.001$). The overall significance is therefore driven largely by the effect of prediction rather than by a consistent representational difference.


 Taken together, the subject-wise results do not show a uniform advantage for either representation: the relative performance of the two pipelines varies across held-out groups in a manner associated with the prevalence of partial body visibility, and the aggregate difference is driven substantially by the effect of prediction rather than by a consistent representational gap. The following section examines the two representations under a controlled visibility manipulation on UR Fall dataset, in which body visibility is the axis of the train/test split.



\subsubsection{Occlusion-Based Generalization}


\begin{table}[h]
\caption{Comparison of classification performance (mean ± standard deviation across 5 validation folds for a representative seed) using pipelines with supervised and unsupervised keypoints on the UR Fall dataset for occlusion-based generalization.}
\label{tab:occlusion}
\centering
\setlength{\tabcolsep}{12pt}
\scriptsize
\begin{tabular}{@{}lcccccccc@{}}
\toprule
& \multicolumn{2}{c}{Accuracy (\%)} 
& \multicolumn{2}{c}{Precision (\%)} 
& \multicolumn{2}{c}{Recall (\%)} 
& \multicolumn{2}{c}{F1-Score (\%)} \\
\cmidrule(lr){2-3} \cmidrule(lr){4-5} \cmidrule(lr){6-7} \cmidrule(lr){8-9}
Prediction type 
& Unsup.\footnotemark[1] & Sup.\footnotemark[1]
& Unsup. & Sup. 
& Unsup. & Sup. 
& Unsup. & Sup. \\
\midrule
Without      & \textbf{72.2}±10.5 & 65.6±2.2 & 66.8±7.9 & \textbf{79.6}±11.4 & \textbf{88.9}±9.9 & 46.7±14.7 & \textbf{76.3}±8.7 & 56±10.4 \\
With      & \textbf{74.4}±9.7 & 50±5 & \textbf{75.9}±15.2 & 50±12.3 & \textbf{77.8}±7 & 24.4±8.3 & \textbf{75.8}±7.8 & 32.2±9.3 \\
\bottomrule
\end{tabular}
\footnotetext[1]{Sup. refers to supervised and Unsup. refers to unsupervised.}
\end{table}

As in the random-split and subject-wise experiments, the occlusion-based experiment was repeated with three random seeds under both prediction settings, yielding paired unsupervised-supervised F1 comparisons. The full classification performance on UR Fall dataset for occlusion-based generalization for a representative seed under both prediction settings and the six paired mean F1 scores across validation folds (three seeds × two prediction settings) are shown in Table~\ref{tab:occlusion} and Table \ref{tab:f1_pairs_ur} respectively. This evaluation uses a small, condition-controlled test set of 9 fall sequences and 9 ADL sequences.  Table~\ref{tab:occlusion} demonstrates that unsupervised keypoints consistently outperform supervised keypoints across both prediction settings, with a substantially higher F1 score for unsupervised keypoints, confirming that the overall detection quality is meaningfully better. The difference is most pronounced in detection sensitivity (recall), which is the clinically critical quantity in a fall-monitoring context, since a missed fall carries greater risk than a false alarm. Without prediction, the unsupervised pipeline detected on average approximately 8 of the 9 fall sequences (recall 88.9\%), whereas the supervised pipeline detected only about 4-5 of 9 (recall 46.7\%); that is, it failed to respond to roughly half of all fall events. These results are consistent with our hypothesis that supervised keypoints degrade under occlusion because they are constrained to localize predefined anatomical landmarks that may be partially or fully absent, while unsupervised keypoints, not being tied to fixed anatomical locations, can adapt their spatial anchors to whatever body structure remains visible.

\textbf{One-sided Wilcoxon signed-rank test:} To evaluate whether the pipeline using the unsupervised keypoint detector outperformed the pipeline using the supervised keypoint detector under occlusion-based generalization, a paired one-sided Wilcoxon signed-rank test was applied on the six paired mean F1 scores. The Wilcoxon signed-rank test was chosen as a non-parametric test appropriate for small samples, as it makes no normality assumption on the paired differences.

The one-sided test was chosen because our hypothesis was directional: the unsupervised keypoint detector was expected to perform better under occlusion-based generalization because of anatomical independence. The results showed that the unsupervised keypoint detector achieved significantly higher F1 scores than the supervised keypoint detector, ($W^+$ = 21.0), (p = 0.0156 $<$ 0.05). We note that for n = 6 the smallest attainable one-sided p value is 0.0156; a significant result at this sample size therefore corresponds to all six pairs favoring the same pipeline. The median paired improvement was 28.8 F1 points; considered separately, 21.9 without prediction and 39.9 with prediction, indicating that the unsupervised keypoint detector provided a consistent performance advantage in the occlusion-based generalization setting.

\begin{figure}
\centering
\includegraphics[width=\textwidth]{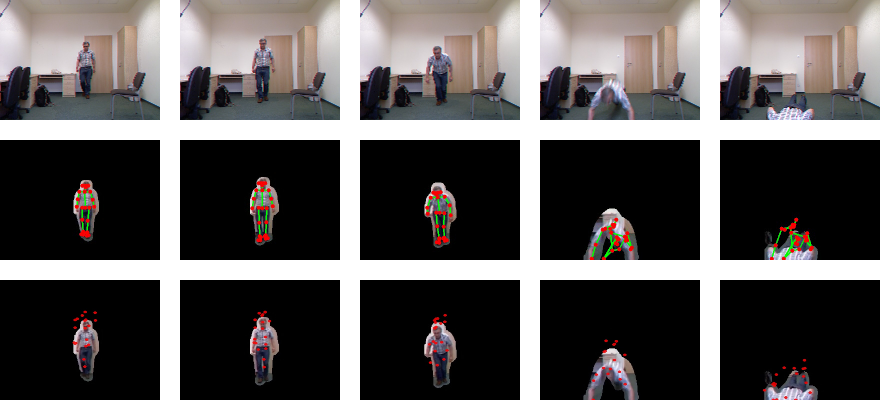}
\caption{Supervised and unsupervised keypoints overlaid on video frames from the UR Fall dataset. Row 1 shows the original video frames, Row 2 shows frames with supervised keypoints on person-centric region, and Row 3 shows frames with unsupervised keypoints on person-centric region.}\label{fig_3}
\end{figure}

\begin{table}[h]
\caption{Paired mean F1 scores across validation folds for three random seeds on the UR Fall dataset for occlusion-based generalization. The first and second value of the pair refer to the mean F1 score obtained from using the pipeline with unsupervised keypoint detector and supervised keypoint detector respectively.}
\label{tab:f1_pairs_ur}
\centering
\begin{tabular*}{\textwidth}{@{\extracolsep{\fill}}lccc@{}}
\toprule
Prediction type
& \textbf{Seed\#1}
& \textbf{Seed\#2} 
& \textbf{Seed\#3} \\
\midrule
Without      & (\textbf{76.3}, 56) & (\textbf{76.5}, 52.1) & (\textbf{73.9}, 52) \\
With      & (\textbf{75.8}, 32.3) & (\textbf{74.3}, 34.4) & (\textbf{69.5}, 36.3) \\
\bottomrule
\end{tabular*}
\end{table}
Figure \ref{fig_3} shows qualitative examples of supervised and unsupervised keypoints on UR Fall video frames with person-centric region.

\section{Discussion}

\subsection{Advantages of Unsupervised Keypoints Over Supervised Keypoints in Fall Detection}

The key question of this study is not which detector is universally superior, but which motion representation remains most reliable under the visibility, variability, and deployment constraints that characterize real clinical monitoring. Our findings suggest that the unsupervised keypoints preserved the temporal trend of the movement more reliably, especially for distinguishing broad posture transitions during falls under realistic deployment conditions.

Unsupervised keypoints offer several practical advantages over supervised keypoints for fall detection. 
Rather than relying on consistent localization of predefined anatomical joints, the unsupervised detector tends to capture salient motion structure and overall body configuration. Because of this, it can remain effective even when image quality is degraded by partial occlusion, or moderate changes in viewpoint. In contrast, supervised keypoint detectors depend on consistent localization of predefined anatomical joints. While this can be beneficial when the image quality is high and body pose is clearly visible, it also makes the representation vulnerable to missing-joint errors during posture change, self-occlusion, or partial exit from the camera field of view. These issues are particularly important in fall scenarios, where the body may rotate quickly, overlap with furniture, bend sharply, or move partially out of view. Once some joints are missed, the downstream classifier receives incomplete temporal information, which can degrade both precision and recall.


\textbf{Occlusion Robustness Through Anatomical Independence}: The advantage of unsupervised keypoints under occlusion stems from their anatomical independence. 
Detectors that are trained to localize a predefined set of anatomical joints; the class to which OpenPose belongs are constrained to report those specific landmarks; when the landmarks are occluded or outside the camera frame, such a detector tends to produce degraded or hallucinated detections at anatomically expected but visually absent locations. This degrades the downstream representation precisely when accurate structural encoding is most critical for distinguishing a fall from normal activity. Unsupervised keypoints, having no such anatomical constraint, naturally gravitate toward whatever salient visual structure remains visible, providing the downstream classifier with meaningful spatial anchors derived from partial observations rather than failing on absent ones. Under clean conditions with full body visibility, however, this flexibility offers no particular advantage; supervised keypoints provide a strong and consistent anatomical prior that benefits classification when all landmarks are reliably detectable. The relative advantage of each approach is therefore condition-dependent: in our experiments, supervised keypoints are favored under clean full-body visibility, while unsupervised keypoints are favored when partial occlusion renders predefined anatomical landmarks unreliable.

At the same time, the flexibility of unsupervised keypoints can also be a limitation when ADL motions are highly varied and include rapid posture changes such as repeated sitting, standing, bending, or lying down. In such cases, the motion representation may separate complex non-fall transitions less sharply, which can lead to false alarms. Thus, the main strength of unsupervised keypoints is not that they are universally superior in all cases, but that they provide a more robust and deployable representation for fall detection in challenging real-world conditions.

\textbf{Privacy-preserving monitoring settings:} Another important advantage is that unsupervised keypoints appear to be better aligned with privacy-preserving monitoring settings. 
In hospital rooms, assisted-living facilities, or home monitoring, 
privacy considerations often discourage the use of high-detail visual streams. Because supervised detectors usually rely more heavily on clear visual anatomy, their performance may deteriorate under these conditions. Unsupervised keypoints, by contrast, can still encode coarse but meaningful motion patterns without depending strongly on fine-grained body-joint visibility. This makes them attractive for systems that prioritize privacy, lightweight computation, and bandwidth efficiency.


\textbf{When supervised keypoints are needed for fall detection:} While unsupervised keypoints offer important advantages for robust and deployment-oriented fall detection, supervised keypoints remain valuable in several situations. A key benefit of supervised keypoints is that they correspond to predefined anatomical joints, such as the head, shoulders, elbows, hips, knees, and ankles. Because of this explicit semantic meaning, they provide a representation that is easier to interpret clinically and biomechanically. In some fall-detection settings, it is not enough to determine only whether a fall occurred; clinicians may also want to understand how the fall occurred, which body parts were involved, whether the impact appears to involve the head or hip, or whether the movement pattern suggests instability, weakness, or a specific gait-related issue. Supervised keypoints are better suited for such analyses because the detected points remain tied to recognizable body landmarks.

Another setting where supervised keypoints may be important is when the downstream system must support explainability and standardized reporting. In healthcare applications, practitioners may have greater trust in a system if its output can be linked to familiar concepts such as “the trunk angle changed rapidly,” “the knees buckled,” or “the head descended below hip level.” Such explanations are easier to construct from supervised anatomical joints than from unsupervised keypoints that do not have fixed semantic identities.

Thus, the choice between supervised and unsupervised keypoints should depend on the application goal. If the priority is robust fall detection under practical deployment constraints, unsupervised keypoints may offer a better balance of resilience and deployability. However, if the application requires anatomical interpretability, fine-grained motion assessment, clinical explainability, or biomechanical analysis, supervised keypoints remain important despite their greater sensitivity to occlusion. 

\vspace{-0.08cm}

\subsection{Forecasting and Bandwidth-Reduction Analysis}

An important advantage of the proposed pipeline is that the keypoint predictor enables bandwidth savings by reducing how often full motion observations need to be transmitted. In a conventional video-monitoring system, maintaining continuous awareness of patient activity may require frequent transmission of raw frames or continuously updated motion data, both of which can become costly in bandwidth-constrained settings. In contrast, our pipeline first extracts a compact keypoint-based motion representation and then uses a forecasting module to predict future keypoints over the next set of frames. This means that the system does not need to transmit every original motion observation at every time step. Instead, a limited amount of observed motion can be used to generate a short-term forecast, and the downstream classifier can operate on either the original keypoints, the predicted keypoints, or a combination of both.

This forecasting-based design is especially important for remote patient monitoring in rural or underserved settings, where continuous high-rate data transmission may be impractical. By predicting likely near-future motion locally, the device can reduce communication burden while still preserving clinically meaningful temporal information for fall detection. In effect, the keypoint predictor acts as a temporal compression mechanism: rather than sending all keypoint sequences frame by frame, the system can send a subset of observed motion information and use prediction to fill in the short-term evolution of the subject's movement. This can substantially reduce communication costs while maintaining situational awareness.

A further implication of the prediction module concerns keypoints that fail to arrive at the monitoring end. In practical deployments, some keypoint frames may be unavailable due to temporary communication dropouts,  noise, occlusion, or unstable detection. In such cases, the forecasting model can help bridge short gaps in the motion sequence by providing plausible estimates of the missing future trajectory. Although predicted keypoints are not a replacement for true observations, they can preserve temporal continuity and reduce the impact of short missing segments on downstream fall classification. 
\vspace{-0.08cm}
\subsection{Application in Remote Patient Monitoring}

The implications of this work for healthcare extend beyond fall classification accuracy. The proposed pipeline is designed around a practical healthcare objective: enabling continuous fall-aware monitoring in settings where privacy, bandwidth, staffing, and deployment cost are major constraints. By converting video into compact motion keypoints and optionally forecasting future motion locally, the system supports a form of monitoring that is more compatible with real-world care delivery than continuous transmission of raw video. This makes the framework relevant not only as a computer-vision method, but also as a digital-health tool for scalable patient safety monitoring.

\begin{figure}
\centering
\includegraphics[width=\textwidth]{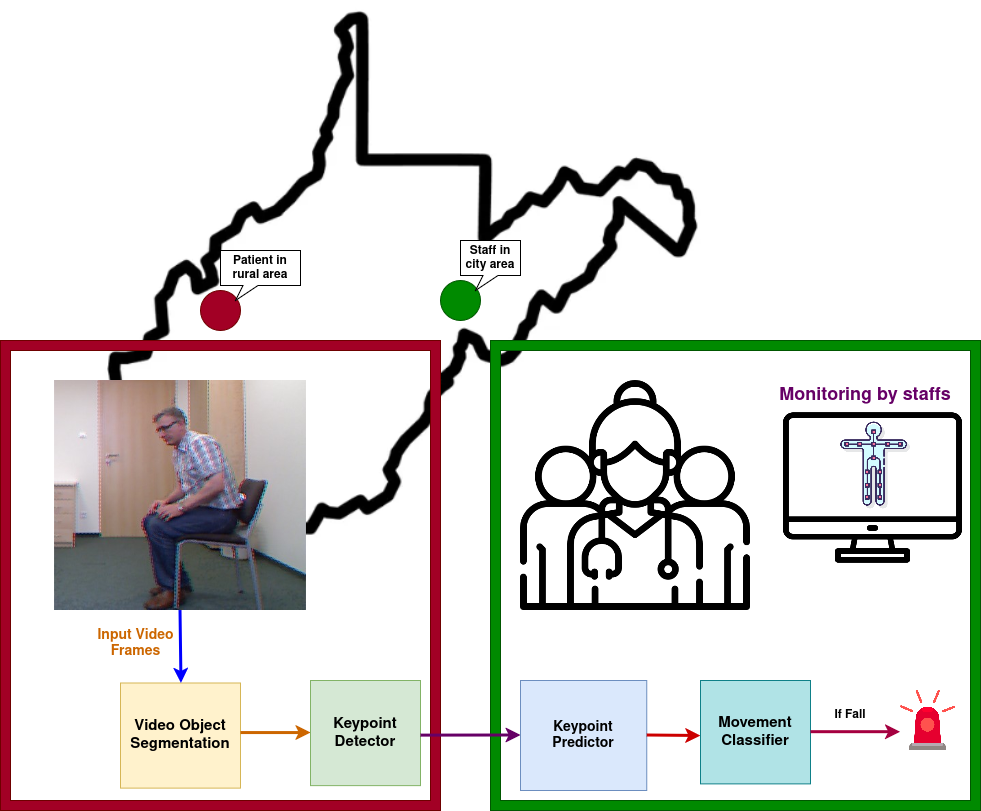}
\caption{Remote patient monitoring using the proposed pipeline}\label{fig_4}
\end{figure}


One important application is remote patient monitoring for rural care delivery as shown in Figure \ref{fig_4}. In regions such as rural West Virginia, older adults may live far from hospitals and specialty care centers, while broadband connectivity may be limited or inconsistent. In such settings, a home camera system with a keypoint detector could extract motion keypoints locally and transmit only compact motion summaries to a hospital or monitoring center instead of continuously streaming full video. After a fall is detected, the staff can route the alert to the nearest family member, neighbor/contact person, home health nurse, or EMS if no one can respond fast enough. The addition of forecasting further strengthens this model by reducing how often observed motion data must be transmitted, since short-term future motion can be estimated locally. This type of bandwidth-aware design is especially important for home-to-hospital monitoring, where the goal is to maintain awareness of patient safety without imposing the communication cost of continuous video streaming. While end-device classification with transmission of only the final alarm can minimize bandwidth, such an approach provides limited interpretability and little support for secondary monitoring tasks. By contrast, transmitting compact motion representations preserves temporal information that could additionally support alert verification and coarse longitudinal indicators of activity level.

The framework is also relevant to aging-in-place and assisted-living support. Many older adults prefer to remain in their homes rather than move into institutional care, but living alone increases the risk that a fall may go unnoticed for a prolonged period. A privacy-aware monitoring system based on compact motion representation could provide continuous room-level fall awareness without requiring identity-rich video to be transmitted at all times. In this sense, the approach supports a middle ground between no monitoring and fully intrusive video surveillance. Such a design is well aligned with reducing time spent unattended after a fall.

More broadly, the work suggests that privacy-preserving motion abstraction may be a useful design strategy for remote fall monitoring. Many healthcare applications require longitudinal sensing in private environments such as bedrooms, patient rooms, or recovery spaces. In such settings, raw RGB video raises substantial privacy and acceptability concerns. By operating on keypoint trajectories rather than identity-rich frames, the proposed framework reduces the exposure of sensitive visual information while preserving motion cues needed for safety monitoring. This makes the approach more compatible with clinical adoption, patient trust, and scalable deployment in sensitive care environments.

Taken together, these scenarios suggest that the contribution of this work is not limited to fall detection as a narrow classification task. Rather, it points toward a broader patient monitoring model in which edge-based motion extraction, forecasting, and selective alerting can support safer and more efficient care across the home and assisted-living facility. In that sense, the pipeline should be viewed as a component of a future monitoring ecosystem that is predictive, privacy-aware, and operationally aligned with real clinical workflows. 

\subsection{Compute Advantages for Practical Deployment}

In the proposed pipeline, a YOLOv8-m segmentation model is first executed on the patient-side device to isolate the person from the scene in each video frame, and the segmented output is then passed to the keypoint detector. A major advantage of the pipeline is the substantially lower compute requirement of the unsupervised keypoint detector compared with the supervised alternative for real-time operation. Using the relation

\[
\text{Trillion Operations per second (TOPS)}=\frac{\text{FLOPs per output frame}\times \text{fps}}{u \times 10^{12}},
\]

\noindent at 30 fps (frames per second), assuming an average device utilization $u = 60\%$ \cite{gao2024empirical}, the estimated compute required for the segmentation model is $\sim$5.5 TOPS \cite{ultralytics_yolov8_docs} and for the OpenPose-based supervised detector the estimated compute demand is approximately 2.0--6.8 TOPS \cite{osokin2018real}, whereas the unsupervised keypoint detector we have used requires only about 0.026--0.1 TOPS (per-layer FLOP counts are given in Appendix Tables A2 and A3).

Thus, although both versions of the pipeline share the same segmentation cost, the choice of keypoint detector still creates a very large compute difference in the overall system. In particular, once segmentation is fixed, the unsupervised keypoint stage adds only a very small extra compute burden relative to the supervised OpenPose-based alternative. At the same frame rate, the unsupervised detector requires roughly two orders of magnitude lower compute than the supervised alternative. This gap is highly significant for practical healthcare monitoring, where the perception front-end may need to run continuously and in real time.

This compute reduction has direct implications for the applications considered in this work. In remote patient monitoring for rural care delivery, lower TOPS demand makes the unsupervised detector much more suitable for deployment on home-side edge devices, where power, hardware capability, and thermal budget may all be limited. A lightweight detector makes it feasible to extract motion information locally and transmit only compact keypoint summaries or forecasted motion features to a hospital or monitoring center, instead of transmitting continuous raw video. In this way, reduced computation complements reduced bandwidth, making the system more practical for home-to-hospital monitoring.

The compute advantage is also important for aging-in-place and assisted-living settings, where monitoring may need to run continuously over long periods without expensive local hardware. A detector requiring 2.0--6.8 TOPS at 30 fps can become difficult to scale across multiple residents, particularly on embedded platforms. In contrast, the much smaller TOPS requirement of the unsupervised detector makes large-scale, always-on monitoring more realistic. This is especially relevant in privacy-aware systems, where the goal is to keep processing local and avoid continuous streaming of identity-rich visual data.


Maximum bandwidth consumption for sending and receiving high quality video is $V_b = 1000$ to $1500$ kbps \cite{webex_min_bandwidth_meetings}.
The bitrate estimate using FP32 data representation format at 30 fps for supervised and unsupervised keypoints can be obtained from $R_{model}(kbps) = D \times 32 \times 30 / 1000$ where D is the keypoint dimension. After using keypoint prediction to forecast half of the total number of frames, $R_{model}$ is $0.48D$. Since the OpenPose model provides 25 body keypoints and we have chosen 20 as the number of keypoints for the unsupervised keypoint detector, $D=(25\times3)=75$ for supervised keypoints and $D=(20\times3)=60$ for unsupervised keypoints, which yield the rates to 36 and 28.8 kbps respectively. To calculate bandwidth saving, we use $V_b/R_{model}$ and obtain $27.78$ to $41.67 \times$ bandwidth saving for supervised keypoints and $34.7$ to $52.1 \times$ bandwidth saving for unsupervised keypoints compared to the practical baseline of $1000$ to $1500$ kbps. 

This TOPS comparison and bandwidth saving values highlight that the benefit of the unsupervised detector is not only predictive performance, but also deployability. A compute-efficient keypoint extractor lowers barriers related to hardware cost, power consumption, thermal constraints, and scaling across many monitored spaces. For the fall-aware remote monitoring applications targeted in this work, these properties make the unsupervised keypoint detector a more practical foundation for privacy-preserving, bandwidth-aware, and always-on fall monitoring.

\vspace{-0.08cm}
\section{Conclusions and Future work}

This work investigated whether unsupervised motion keypoints provide a more suitable representation than supervised pose estimation for fall detection under realistic deployment conditions. We proposed a privacy-preserving remote fall-monitoring pipeline based on unsupervised keypoint extraction, predictive temporal modeling, and lightweight classification, adopting a distributed edge-cloud architecture that supports bandwidth reduction through intermittent keypoint transmission.
Our central finding is that the relative advantage of each representation is condition-dependent. Conventional random-split protocols are insufficient to reveal meaningful representational differences, whereas in subject-disjoint evaluation neither representation holds a uniform advantage, with relative performance varying across held-out subjects in a manner associated with body visibility. Under occlusion-based out-of-distribution evaluation, unsupervised keypoints substantially outperform supervised keypoints, and this advantage persists in the bandwidth-constrained predictive setting. These results support the argument that anatomical independence is the source of occlusion robustness in unsupervised representations. Beyond detection performance, the unsupervised detector requires roughly two orders of magnitude less computation than the supervised alternative, reinforcing its suitability for privacy-preserving, always-on monitoring on edge hardware. While more recent supervised pose estimators differ in architecture, they share a reliance on predefined anatomical structure; characterizing their behavior and their accuracy-compute trade-offs under the visual conditions studied here is a natural extension of this work.

The findings suggest that for fall monitoring under practical deployment constraints, the choice of keypoint representation should be guided by the expected visual conditions rather than benchmark performance alone. Although the proposed framework shows strong promise for privacy-preserving and deployment-oriented fall monitoring, several important directions remain for future work. First, our results indicate that the unsupervised keypoint representation is not uniformly optimal across all motion types. 
Future work should therefore focus on improving the robustness of the learned representation under varying motion regimes. 

A second direction is to extend the framework beyond video-only motion cues by incorporating multimodal data. 
Examples include inertial measurements from wearable accelerometers or gyroscopes, depth signals, pressure sensors near the bed or floor, audio events, vital signs, or electronic health record variables such as medication status, prior fall history, and mobility risk scores. Integrating such modalities could improve robustness in cases where vision alone is ambiguous, for example under occlusion, or unusually rapid motion transitions.

Another important direction for future work is validation on genuine falls from older adults. Extending the evaluation to spontaneous falls in the target population is a next step toward clinical deployment.








\section*{Appendix}
\renewcommand{\thetable}{A\arabic{table}}
\setcounter{table}{0}

Models were trained on NVIDIA RTX 3090 and NVIDIA RTX A6000. Hyperparameters that are used to train the proposed pipeline are given in Table \ref{hyp} and layer-by-layer FLOP estimate of the unsupervised keypoint detector are given in Tables \ref{tab:flops_64} and \ref{tab:flops_128}. 

\begin{table*}[t]
\centering
\caption{Hyperparameters for training keypoint detector, predictor and classifier}
\label{hyp}
\begin{tabular}{p{5cm} p{3cm} p{3cm}}
\toprule
\textbf{Parameter name} & \textbf{UR Fall} & \textbf{Human Fall} \\
\midrule
Input resolution & 64x64 & 128x128 \\
\midrule
Learning rate & 0.001 & 0.001 \\
\midrule
Observed steps & 12 & 12\\
\midrule
Predicted steps & 12 & 12 \\
\midrule
No. of keypoints & 20 & 20 \\
\midrule
Heatmap width & 16 & 32 \\
\midrule
Heatmap regularization & 5 & 5 \\
\midrule
Keypoint sparsity scale & 0.001 & 0.001 \\
\midrule
Keypoint separation scale & 50 & 50 \\
\midrule
Separation loss width & 0.1 & 0.1 \\
\midrule
Keypoint blob width & 1.5 & 3 \\
\midrule
Latent code size & 16 & 16 \\
\midrule
KL loss scale & 0.0001 & 0.0001 \\
\midrule
Prior net size & 128 & 128 \\
\midrule
Posterior net size & 128 & 128 \\
\midrule
No. of RNN units \\ used in keypoint predictor & 512 & 512 \\
\midrule
Classifier loss scale & 0.1 - 0.5 & 0.1 - 0.5 \\
\midrule
No. of RNN units \\ used in classifier & 128 & 128 \\
\bottomrule
\end{tabular}
\end{table*}

\begin{table*}[t]
\centering
\caption{Layer-by-layer FLOP estimate of the unsupervised keypoint detector
(image encoder and final $1\times1$ heatmap layer) at $64\times64$ input
(UR Fall setting, \texttt{heatmap\_width}~$=16$). MACs $=$
$H_{\text{out}}\!\times\!W_{\text{out}}\!\times\!C_{\text{out}}\!\times\!C_{\text{in}}\!\times\!k^2$;
FLOPs $=2\times$MACs. Pointwise operations (coordinate channels, normalization,
softplus) are negligible and omitted.}
\label{tab:flops_64}
\begin{tabular}{lccccrr}
\toprule
Layer & Output & $C_{\text{in}}$ & $C_{\text{out}}$ & $k$ & MMACs & MFLOPs \\
\midrule
Input conv (stride 1)    & $64\times64$ & 3   & 32  & 3 & 3.54   & 7.08 \\
Full-res conv 1          & $64\times64$ & 32  & 32  & 3 & 37.75  & 75.50 \\
Full-res conv 2          & $64\times64$ & 32  & 32  & 3 & 37.75  & 75.50 \\
Downsample 1 (stride 2)  & $32\times32$ & 32  & 64  & 3 & 18.87  & 37.75 \\
\quad conv 1 @ $32\times32$ & $32\times32$ & 64  & 64  & 3 & 37.75  & 75.50 \\
\quad conv 2 @ $32\times32$ & $32\times32$ & 64  & 64  & 3 & 37.75  & 75.50 \\
Downsample 2 (stride 2)  & $16\times16$ & 64  & 128 & 3 & 18.87  & 37.75 \\
\quad conv 1 @ $16\times16$ & $16\times16$ & 128 & 128 & 3 & 37.75  & 75.50 \\
\quad conv 2 @ $16\times16$ & $16\times16$ & 128 & 128 & 3 & 37.75  & 75.50 \\
Heatmap $1\times1$ conv   & $16\times16$ & 128 & 20  & 1 & 0.66   & 1.31 \\
\midrule
\textbf{Total per frame} & & & & & \textbf{268.4} & \textbf{536.9} \\
\bottomrule
\end{tabular}
\end{table*}

\begin{table*}[t]
\centering
\caption{Layer-by-layer FLOP estimate of the unsupervised keypoint detector
at $128\times128$ input (Human Fall setting, \texttt{heatmap\_width}~$=32$).
Same convention as Table~\ref{tab:flops_64}.}
\label{tab:flops_128}
\begin{tabular}{lccccrr}
\toprule
Layer & Output & $C_{\text{in}}$ & $C_{\text{out}}$ & $k$ & MMACs & MFLOPs \\
\midrule
Input conv (stride 1)     & $128\times128$ & 3   & 32  & 3 & 14.16   & 28.31 \\
Full-res conv 1           & $128\times128$ & 32  & 32  & 3 & 150.99  & 301.99 \\
Full-res conv 2           & $128\times128$ & 32  & 32  & 3 & 150.99  & 301.99 \\
Downsample 1 (stride 2)   & $64\times64$   & 32  & 64  & 3 & 75.50   & 150.99 \\
\quad conv 1 @ $64\times64$  & $64\times64$   & 64  & 64  & 3 & 150.99  & 301.99 \\
\quad conv 2 @ $64\times64$  & $64\times64$   & 64  & 64  & 3 & 150.99  & 301.99 \\
Downsample 2 (stride 2)   & $32\times32$   & 64  & 128 & 3 & 75.50   & 150.99 \\
\quad conv 1 @ $32\times32$  & $32\times32$   & 128 & 128 & 3 & 150.99  & 301.99 \\
\quad conv 2 @ $32\times32$  & $32\times32$   & 128 & 128 & 3 & 150.99  & 301.99 \\
Heatmap $1\times1$ conv    & $32\times32$   & 128 & 20  & 1 & 2.62    & 5.24 \\
\midrule
\textbf{Total per frame}  & & & & & \textbf{1073.7} & \textbf{2147.5} \\
\bottomrule
\end{tabular}
\end{table*}

\section*{Acknowledgements}

The authors would like to acknowledge the Pacific Research Platform, NSF Project
ACI-1541349, and Larry Smarr (PI, Calit2 at UCSD) for providing the computing
infrastructure used in this project. Large language model (LLM) assistance was used in drafting explanatory text for selected findings. All content was reviewed, verified, and edited by the authors, who take full responsibility for the accuracy and integrity of the work. The figures were created by the authors using draw.io (diagrams.net), and some icons used within the figures were obtained from Flaticon (flaticon.com).





\section*{Conflict of interest}

The authors declare that the research was conducted in the absence of any commercial or financial relationships that could be construed as a potential conflict of interest.

\section*{Author contributions}

T.H.: Methodology, Software, Validation, Visualization, Formal Analysis, Original Draft; J.K.: Data Preprocessing, Validation, Visualization; S.M.: Discussion; M.A.A.M.: Conceptualization, Clinical and Health Systems Methodology, Discussion, Supervision, Research Funding; S.D.: Conceptualization, Computational and Algorithmic Methodology, Discussion, Supervision, Computational Resources, Research Funding. All authors contributed to manuscript editing and revision.

\section*{Data Availability}

The datasets analyzed in this study are publicly available: the UR Fall Detection dataset \cite{kwolek2014human} and the Human Fall Detection/ GMDCSA-24 dataset \cite{alam2024gmdcsa}. The occlusion-based train/test partition of the UR Fall dataset was constructed by manual visual inspection; the exact assignment of videos to the training and test sets is provided in our code repository (https://github.com/Tasmiah1408028/Unsupervised-Keypoints-for-Real-Time-Fall-Detection) to support reproducibility. The subject-wise partitions of the Human Fall dataset follow a leave-one-subject-out protocol determined directly by the dataset's subject labels, as described in Section \ref{sec:4.4.2}. Some generated video samples that represent the findings of this research are provided at: https://github.com/Tasmiah1408028/Videos-of-Real-time-Fall-Detection.



\noindent

\bigskip

\bigskip\noindent

\bigskip\noindent

\bigskip\noindent

\bigskip\noindent
\bibliographystyle{unsrt}
\bibliography{sn-bibliography}

\end{document}